# ROS Regression: Integrating Regularization with Optimal Scaling Regression

Jacqueline J. Meulman* and Anita J. van der Kooij**

Leiden University

In this paper we combine two important extensions of ordinary least squares regression: regularization and optimal scaling. Optimal scaling (sometimes also called optimal scoring) has originally been developed for categorical data, and the process finds quantifications for the categories that are optimal for the regression model in the sense that they maximize the multiple correlation. Although the optimal scaling method was developed initially for variables with a limited number of categories, optimal transformations of continuous variables are a special case. We will consider a variety of transformation types; typically we use step functions for categorical variables, and smooth (spline) functions for continuous variables. Both types of functions can be restricted to be monotonic, preserving the ordinal information in the data. In addition to optimal scaling, three popular regularization methods will be considered: Ridge regression, the Lasso, and the Elastic Net. The resulting method will be called ROS Regression (Regularized Optimal Scaling Regression. We will show that the basic OS algorithm provides straightforward and efficient estimation of the regularized regression coefficients, automatically gives the Group Lasso and Blockwise Sparse Regression, and extends them with monotonicity properties. We will also show that Optimal Scaling linearizes nonlinear relationships between predictors and outcome, and improves upon the condition of the predictor correlation matrix, increasing (on average) the conditional independence of the predictors. Alternative options for regularization of either regression coefficients or category quantifications are mentioned. Extended examples are provided.

*Jacqueline J. Meulman is Professor, Mathematical Institute, Leiden University, P.O. Box 9512, 2300 RA Leiden, The Netherlands (E-mail: jmeulman@math.leidenuniv.nl), and Visiting Professor, Department of Statistics, Stanford University (jmeulman@stanford.edu).

**Anita J. van der Kooij is Senior Researcher, Institute of Psychology, Methodology and Statistics Division, Leiden University, P.O. Box 9555, 2300 RB Leiden, The Netherlands (E-mail: kooij@fsws.leidenuniv.nl).

The current version of this paper was written while the first author was in the Stanford Statistics Department.

# 1. INTRODUCTION

Multiple regression investigates the relationship between a response (outcome) variable and a set of predictor variables, and can be used to estimate a model for predicting future responses. For the first goal the complexity of the model is of primary interest, while for the second goal the prediction accuracy of the model is more important. Ordinary least squares (OLS) regression is known for often not performing well with respect to both model complexity and prediction accuracy. While regularization addresses the prediction accuracy problem, optimal scaling can be seen as a method that addresses the problem of model complexity. At the same time, optimal scaling can deal with categorical predictor and response variables. It is well-known that OLS regression may result in highly variable estimates of the regression coefficients in the presence of collinearity, or when the number of predictors ($P$) is large relative to the number of objects ($N$). Biological and chemical research, sometimes subsumed under the term 'omics', generates data where the number of objects is much smaller than the number of predictors, and this situation demands some form of regularization. Data often require more complex models, but while these may decrease the bias (we are able to fit local structure in the data), they might increase the variance of the estimates too much. It is well-known that high variance gives poor predictions, and the variance may be reduced by increasing the bias. In those cases, we add a penalty term to the loss function that controls the variance of the regression coefficients, hereby decreasing the standard error of the estimates. Over the years, several methods for regularized regression have been developed. Without any claim to be complete, regularization methods began with Ridge regression (Hoerl and Kennard 1970a,b), followed by Bridge regression (Frank and Friedman 1993), the Garotte (Breiman 1995), and the Lasso (Tibshirani 1996), and were followed somewhat later by LARS (Efron, Hastie, Johnstone, and Tibshirani 2004), Pathseeker (Friedman and Popescu 2004), and the Elastic Net (Zou and Hastie 2005). Since then the number of references especially to the Lasso and its extensions has grown exponentially.

The oldest regularization method, Ridge regression, reduces the variability by shrinking the coefficients, resulting in less variance at the cost of usually only a small increase of bias. The coefficients are shrunken towards each other and to zero, but will never become exactly zero. So, when the number of predictors is large, Ridge regression will not provide a sparse model that is easy to interpret. Subset selection, on the other hand, does provide interpretable models, but assumes extreme sparseness. The Lasso was developed by Tibshirani (1996) to improve both prediction accuracy and model interpretability by combining the nice features of Ridge regression and subset selection. Thus, the Lasso reduces the variability of the estimates by shrinking the coefficients, and at the same time produces interpretable models by shrinking some coefficients to exactly zero. The Elastic Net (Zou and Hastie 2005) combines Ridge regression and the Lasso, obtaining sparse models due to the use of a Lasso penalty, and encouraging grouping of variables due to the use of a Ridge penalty. Where the Lasso would only select one variable of the group, the Elastic Net tends to select groups of highly correlated variables together.

The original Lasso algorithm uses a quadratic programming strategy that is complex and computationally demanding; hence it is not feasible for large values of $P$, and moreover, it can not be used when $P > N$. Since the Lasso paper, various less complex and/or more efficient lasso algorithms were proposed. For example, Osborne, Presnell, and Turlach (2000a) developed a homotopy method that can handle $P > N$ predictors, but it is still computationally demanding when $P$ is large. The same method was discussed in Efron et al. 2004 in a different framework, and became known as the LARS-Lasso. These methods provide efficient algorithms to find the entire Lasso regularization path. The "Grafting" algorithm of Perkins, Lacker, and Theiler (2003), the "Pathseeker" algorithm of Friedman and Popescu (2004), and the "boosting" algorithm of Zhao and Yu (2004) are gradient descent algorithms that can deal with $P > N$ predictors in a computationally less demanding way. However, in the $P > N$ case, none of these Lasso algorithms can select more than $N$ predictors. The Elastic Net algorithm that is based on the LARS-Lasso algorithm is capable of selecting more than $N$ predictors due to the use of the additional Ridge penalty. All these methods apply only to linear regression.

In this paper, we show how to implement Ridge, Lasso, and Elastic Net penalties in regression models for categorical data (both nominal and ordinal), as well as in generalized additive models, with nonmonotonic and monotonic transformation of the data. The resulting straightforward estimation of regularized coefficients enables the Lasso to select more than $N$ predictors. The algorithm employs optimal scaling (Gifi 1990; also called optimal scoring in Buja (1990)) to transform the predictors (and sometimes the outcome variable). We usually apply step functions when analyzing categorical data, and use smooth spline functions for continuous data. The number of parameters estimated is controlled by the degree of the spline and the number of internal knots.

The straightforward estimation of regularized coefficients and the ability to select more than $N$ predictors is due to the "one-variable-at-a-time" approach that is the basis of the Optimal Scaling algorithm. The application of the "one-variable-at-a-time" approach in regression was originally used to find transformations of the data, as in De Leeuw, Young, and Takane (1976), Friedman and Stuetzle (1981), Gifi (1990), Breiman and Friedman (1985), Buja, Hastie, and Tibshirani (1989), and Hastie and Tibshirani (1990). In the psychometric literature, the strategy has been called "alternating least squares" or "conditional least squares", in statistics it was labeled "backfitting", following Friedman and Stuetzle (1981). Other terminology found in the literature are "the Gauss-Seidel algorithm", "Newton-Raphson", "Component-wise update", "Block Relaxation", and "Coordinate descent".

A short note about the discovery of the "one-at-a-time" strategy in relation to the Lasso is in order. The very same strategy to find the Lasso solution in linear regression problems was already applied in Fu (1998), who used the name "shooting algorithm". However, the fact that this algorithm worked was not fully appreciated at the time, or not fully understood. For example, Osborne, Presnell, and Turlach (2000b) state that it is not applicable in the $P > N$ case. The same approach was independently re-invented in Daubechies, Defrise, and De Mol (2004) and in the optimal scaling research in Leiden in 2006, as reported in Van der Kooij (2007), where it was shown that the one-variable-at-a-time algorithm made the computation of regularized coefficients for the Lasso and thus also for the Elastic Net trivially simple. Friedman, Hastie, Höfling, and Tibshirani (2007) subsequently showed that the algorithm was also very fast. (For more history, see "A brief history of coordinate descent for the lasso" at www-stat.stanford.edu/~tibs/stat315a/Supplements/fuse.pdf, p. 7.)

The paper is organized as follows. Section 2 gives a description of the basic OS regression approach, describes a number of diagnostics to evaluate the transformations, and gives a small example with two predictors to illustrate the approach. Section 3 deals with the computational details of the OS regression algorithm. Section 4 describes two extended data analyses to demonstrate the use and properties of OS regression using real data sets; in both examples, a variety of different models is fitted and the results are compared through the use of diagnostics and cross-validation. The first data set contains mixed measurement predictors. Section 5 then describes how regularization with Ridge, Lasso, and Elastic Net penalties is incorporated, and discusses selection of the optimal values for the Ridge and Lasso penalties. Section 6 provides various examples. In section 7, we will discuss the methods developed by Yuan and Lin (2006) and Kim, Kim, and Kim (2006) that became known as the "Group Lasso" and "Blockwise Sparse Regression". The latter methods expand categorical variables to blocks of dummy variables, and continuous variables to blocks of basis functions, and apply regularization to these blocks of variables by joint shrinkage of the dummy coefficients. We will show that these methods are equivalent to particular choices of transformations within regularized optimal scaling regression, and subsequently can be extended with transformations that are restricted to be monotonic. In the discussion, we will briefly discuss more general restrictions on the regression coefficients and/or category quantifications by specifying particular bounds.

## 2. OPTIMAL SCALING REGRESSION

### 2.1 The OS regression loss function

In linear regression problems, we have a system consisting of a random "outcome", "response", or "dependent" variable $Y$ and a set of random "explanatory" ,"predictor", or "independent" variables $X = \{X_k\}_{k=1}^{P}$, where $P$ denotes the number of predictors. The problem defines a "training" sample, $\{y_i, \mathbf{x}_i\}_1^N$ of known values for $Y$ and $X$, where $(y_i, \mathbf{x}_i)$ links the predictor variables of the $i$th object with the $i$th value of the outcome variable, and where $i = 1, \ldots, N$. Using the training data, the model can be written as

$$y_i = \beta' \mathbf{x}_i + \varepsilon_i = \sum_{k=1}^{P} \beta_k x_{ik} + \varepsilon_i,$$

or (in vector notation) as

$$\mathbf{y} = \beta_1 \mathbf{x}_1 + \beta_2 \mathbf{x}_2 +, ..., + \beta_P \mathbf{x}_P + \varepsilon, \tag{1}$$

where $\varepsilon = \{\varepsilon_i\}_{i=1}^{N}$, and the linear combination of predictor variables is formed through a set of regression coefficients in $\beta = \{\beta_k\}_{k=1}^{P}$. We assume that the response and predictor variables are standardized, thus there is no need to fit an intercept. It is convenient to write the optimization task in the form of a least squares loss function:

$$L(\beta) = \sum_{i=1}^{N} ||y_i - \beta' \mathbf{x}_i||^2 = ||\mathbf{y} - \sum_{k=1}^{P} \beta_k \mathbf{x}_k||^2, \tag{2}$$

where $||\cdot||^2$ denotes the squared Euclidean norm. Loss function (2) has to be minimized over the vector of coefficients $\beta$, and solving for the optimal $\beta$ will give a maximum correlation between the linear combination of the predictor variables and the outcome variable. The well-known ordinary least squares (OLS) solution for $\beta$ is obtained as (in matrix notation)

$$\hat{\beta} = (\mathbf{X}'\mathbf{X})^{-1}\mathbf{X}'\mathbf{y}, \tag{3}$$

where $(\mathbf{X}'\mathbf{X})^{-1}$ denotes the inverse of the correlation matrix between the predictor variables.

Nonlinear generalizations of the linear regression problem come in different forms: (a) models that are nonlinear in the parameters and (b) models that are linear in the parameters, but include nonlinear transformations of the variables. Examples of the first type are loglinear analysis of contingency tables, logistic regression, and generalized linear models (GLIM/GLM, Nelder and Wedderburn (1972)). Examples of the second type are generalized additive models (Hastie and Tibshirani 1990) and multivariate adaptive regression splines (Friedman 1991). Nonlinear regression with optimal scaling falls in the latter category, and has been extensively explored in the psychometric literature, starting with Kruskal's 1965 nonlinear version of MONANOVA. This approach was followed upon in additive modeling (ADDALS, De Leeuw et al. 1976) and multiple regression (MORALS, Young, De Leeuw, and Takane 1976). The collective work by the Leiden group at the department of Data Theory resulted in Gifi (1990). Winsberg and Ramsay (1980) replaced Kruskal's original monotonic regression approach (that produces step functions) by monotonic regression splines (that produce smooth piecewise polynomial functions); a nice review is given in Ramsay (1988). In the meantime, optimal scaling had entered the mainstream statistical literature in the Breiman and Friedman (1985) paper on Alternating Conditional Expectations (ACE). Finally, regression with optimal scaling became widely available in statistical packages such as SAS (in a procedure called TRANSREG) and in the CATREG algorithm in SPSS Categories (Meulman, Heiser, and SPSS 1998). For the latter algorithm, the loss function is written as

$$L(\beta, \varphi, \vartheta) = ||\vartheta(\mathbf{y}) - \sum_{k=1}^{P} \beta_k \varphi_k(\mathbf{x}_k)||^2. \tag{4}$$

The arguments over which the function has to be minimized are the weights $\beta = \{\beta_k\}_{k=1}^{P}$, the transformation $\vartheta(\mathbf{y})$ of $Y$, and $\varphi$ that stands for functions $\varphi_k(\mathbf{x}_k)$, i.e., the set of nonlinear transformations $\varphi = \{\varphi_k(\mathbf{x}_k)\}_{k=1}^{P}$. The nonlinear transformation process has been denoted by various names in the literature: in psychometrics it was called *optimal scaling* (a term originally coined Bock (1960)), Nishisato (1980, 1994) called it *dual scaling*, Buja (1990) reintroduced the older term *optimal scoring*, and when the predictor variables are all categorical, the term quantification is used

Gifi (1990). Quantification is also one of the key terms in the data analysis framework developed by Hayashi (1952).

In the optimal scaling approach, there is a large emphasis on the analysis of categorical data; we therefore at the outset introduce an $N \times C_k$ indicator matrix $\mathbf{G}_k$ for each categorical predictor $X_k$. The number of different categories in $X_k$ is indicated by $C_k$, and each column of $\mathbf{G}_k = G_k(\mathbf{x}_k)$ shows by 1-0 coding whether or not an object $i$ scores in category $c_k$ of $X_k, c_k = 1, ..., C_k$. For each variable, we search for a vector of quantifications $\mathbf{v}_k$ that minimizes the overall value of the associated loss function, now written as

$$L(\beta, V, \vartheta) = ||\vartheta(\mathbf{y}) - \sum_{k=1}^{K} \beta_k \mathbf{G}_k \mathbf{v}_k||^2, \tag{5}$$

where $V$ represents the super vector of quantifications (for the $P$ variables). Thus, the optimal scaling mechanism first involves the expansion $G_k(\mathbf{x}_k)$ in $\mathbf{G}_k$, followed by a contraction in $\mathbf{G}_k \mathbf{v}_k$, and the result is the transformation $\varphi_k(\mathbf{x}_k)$. Since a continuous variable can be viewed as a variable with $N$ (number of objects) categories, numeric, continuous variables, and categorical, discrete variables, can be dealt with in the same framework. It should be noted from the start that we only use indicator matrix notation in the equations to show how to obtain optimal quantifications. We do not use indicator matrices (that are extremely sparse) in the computations. In an efficient algorithm, matrix multiplications that involve $\mathbf{G}_k$ are replaced by simple additions.

Within the class of nonlinear transformations, we make the following distinctions. We call a quantification *nominal* if we merely maintain the class membership information in the quantified variable $\mathbf{G}_k \mathbf{v}_k$, or equivalently, in the nominal transformation $\varphi_k(\mathbf{x}_k)$; if two objects $i$ and $i'$ belong to the same category of variable $k$, then

$$x_{ik} = x_{i'k} \implies \varphi_k(x_{ik}) = \varphi_k(x_{i'k}). \tag{6}$$

If a categorical predictor variable contains *order information* on the objects, this information can be preserved in the transformation:

$$x_{ik} < x_{i'k} \implies \varphi_k(x_{ik}) \leq \varphi_k(x_{i'k}) \tag{7}$$

restricting the ordinal quantifications in $\mathbf{v}_k$ so that $v_{1_k} \leq ... \leq v_{C_k}$, and we call the transformation *ordinal*. In the latter case, $X_k$ and $\varphi_k(\mathbf{x}_k)$ are related by a monotonic step function. A linear transformation is a further restriction by preserving interval information as well, and amounts to standardizing the original variable. If the original variable is continuous, and we wish to apply less restrictive transformations than linear ones, we need to limit the number of parameters that are fitted in the nonlinear transformation. For instance, we can use regression splines in which the number of parameters is limited by restricting the degree of the spline and the number of interior knots. Alternatively, we could first make a continuous variable discrete with a fixed number of categories (binning), and subsequently apply optimal category quantification, resulting in a step function. The relation between regression splines functions and step functions is given by the fact that they are equivalent when the number of parameters fitted in the spline function is equal to the number of categories that is quantified.

## 2.2 Diagnostics to evaluate the optimal scaling transformations

The overall criterion that is optimized by the optimal scaling transformations is the multiple correlation $r^2$ between a linear combination of transformed predictor variables and the (transformed) outcome, as displayed in (4). An important diagnostic for a single predictor is its 'predictability' from the other predictors, and the values for the so-called conditional independence are given by the inverse of the diagonal elements of the inverse of the correlation matrix. The elements of this $P$-vector will be called *tolerance values*, defined by

$$\text{TOL} = \frac{1}{diag(\mathbf{R}^{-1})}.$$

Optimal scaling transformations for multiple regression will usually increase the average value of TOL over the various predictors. A suitable candidate for a diagnostic for the condition of the correlation matrix for (transformed) predictors is the so-called Log Determinant Divergence. This measures the difference between matrices by the log determinants of those matrices. In OS regression, we measure the divergence of the correlation matrix $\mathbf{R}$ and the identity matrix $\mathbf{I}$, because $\mathbf{I}$ is the correlation matrix when all predictors are completely uncorrelated. The Log Determinant Divergence (DLD) is then written as (adapted from Dhillon 2008),

$$\begin{aligned}
\text{DLD} = \mathbf{D}_{\ell d}(\mathbf{R}, \mathbf{I}) &= tr(\mathbf{R}) - \log \det(\mathbf{R}) - P \\
&= \sum_{k=1}^{P} (\lambda_k(\mathbf{R}) - \log(\lambda_k(\mathbf{R})) - 1) \\
&= -\sum_{k=1}^{P} \log(\lambda_k)
\end{aligned} \quad (8)$$

Note that this is a 'degenerate' version of Stein's loss

$$tr(\hat{\mathbf{\Sigma}}\mathbf{\Sigma}^{-1}) - \log \det(\hat{\mathbf{\Sigma}}\mathbf{\Sigma}^{-1}) - P,$$

where $\hat{\mathbf{\Sigma}}$ is the estimator of $\mathbf{\Sigma}$.) Equation 8 shows that our diagnostic $\mathbf{D}_{\ell d}$ (DLD) boils down to a simple function of the eigenvalues of the correlation matrix between transformed predictors. Optimal scaling transformations for regression will usually decrease the value of $\mathbf{D}_{\ell d}(\mathbf{R}, \mathbf{I})$. A third diagnostic that can be used to evaluate the condition of $\mathbf{R}$ is the value of its smallest eigenvalue (SMEV). If $\mathbf{R}$ is ill-conditioned, the smallest eigenvalue will be small. Optimal scaling transformations will in general increase the value of the smallest eigenvalue.

## 2.3 Example 1: A simple model with two correlated predictors, nonlinearly related with the outcome

Before going into computational details of the optimal scaling algorithm, we first demonstrate OS regression with a small example. The analysis has two predictor variables only, and we sampled $X_1$ and $X_2$ with $N = 1000$ from a multivariate normal distribution with $\rho = .707$ being the population correlation. The outcome variable was constructed as $\mathbf{y} = \exp(\mathbf{x}_1) + |\mathbf{x}_2| + \varepsilon$, where

Table 1. Results for three different regression models with two predictors.

| Transformation | $r^2$ | $\beta_1$(s.e.) | $\beta_2$(s.e.) | EPE(s.e.) | $r(x_1, x_2)$ | SMEV | TOL* | DLD |
|---|---|---|---|---|---|---|---|---|
| 1. $\lin(x_1)$, $\lin(x_2)$ | .379 | .634(.027) | -.027(.034) | .661(.181) | .706 | .294 | .502 | .690 |
| 2. $\lin(x_1)$, $\spl(x_2)$ | .570 | .637(.029) | .438(.020) | .467(.146) | -.050 | .950 | .998 | .002 |
| 3. $\spl(x_1)$, $\spl(x_2)$ | .855 | .851(.031) | .224(.029) | .148(.007) | .214 | .786 | .954 | .047 |

* In regression with two predictors, both obviously have the same value for the conditional independence.

$\varepsilon \sim \mathcal{N}(0,1)$. The correlation between the predictors in the sample is 0.706. The results for three different models are given in Table 1, which gives the results for the regression coefficients $\beta$, the fit $r^2$, the correlation $r(\mathbf{x}_1, \mathbf{x}_2)$, and the conditional independence values (TOL). The standard error of the regression coefficients has been estimated by a bootstrap with 1000 samples, and the expected prediction error (EPE) and its standard error have been estimated by 10-fold cross-validation. When the predictors were transformed, nonmonotonic spline transformations (using second degree polynomials, with three internal knots) were fitted.

Model 1 gives results for simple linear regression. The predictors are highly correlated, regression coefficients $\beta_1$ and $\beta_2$ are very different, and both the fit ($r^2$) and the prediction accuracy (EPE) are rather poor. Because we only have two predictors, the smallest eigenvalue equals $1 - |r(x_1, x_2)|$. Because $\beta_2$ is very small, we transform $\mathbf{x}_2$, keeping $\mathbf{x}_1$ fixed (model 2); we observe that compared to model 1, the dependence among the predictors becomes minimal (the correlation between the predictors is now $-.050$) and the conditional independence (tolerance) is close to maximal (.998). The $r^2$ increases, as well as the regression coefficient $\beta_2$; the expected prediction error decreases. If we allow both predictors to be transformed (model 3), both $r^2$ and $\beta_1$ increase compared to model 2, while the tolerance values and $\beta_2$ decrease. The expected error rate is smallest for model 3, and compared to model 1, the overall improvement is obvious. Figure 1 shows the partial residual plots, with the partial residual plotted versus predictor $k$. (For example, the plot in the upper left panel depicts $\mathbf{u}_1 = \mathbf{y} - \beta_2 \mathbf{x}_2$ on the vertical axis versus $\mathbf{x}_1$ on the horizontal axis.) These partial residual plots are given for both the original predictors $\mathbf{x}_1$ and $\mathbf{x}_2$ in the left panels, as well as for the transformed predictors $\varphi_1(\mathbf{x}_1)$ and $\varphi_2(\mathbf{x}_2)$ in the right panels. We observe that the transformations $\varphi_1(\mathbf{x}_1)$ and $\varphi_2(\mathbf{x}_2)$, shown in the left middle panels, are a nonlinear fit to the scatter in the partial residual plots in the left panels. The regression between the transformed predictors and the partial residuals in the right middle panels has been linearized, as is seen from the independently fitted smoothing splines (right panels). These functions are fitted to inspect whether the choice of transformation has been appropriate. If not, the plots on the far right hand side would indicate this by showing a nonlinear curve, implying there is still nonlinearity remaining after transformation.

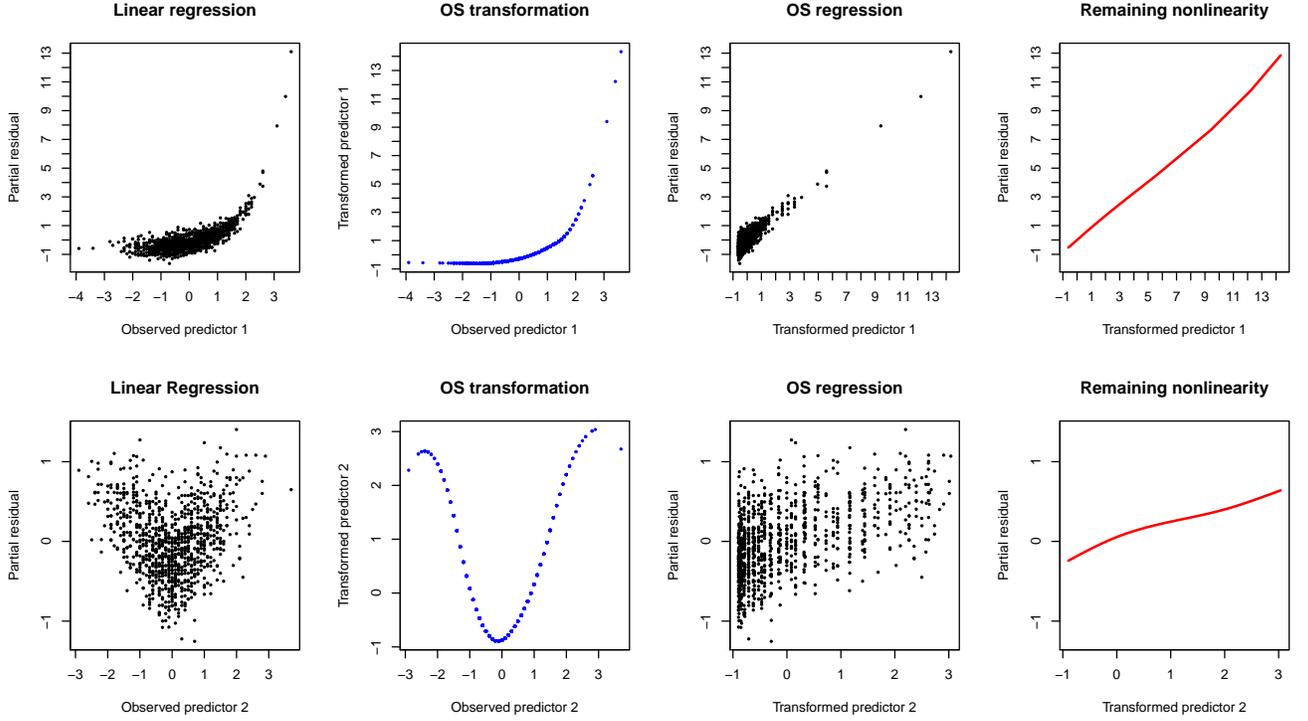

*Figure 1. Two predictors: scatter plots of partial residuals versus observed predictors (left panels) and versus transformed predictors (right middle panels). Transformations $\varphi_1(\mathbf{x}_1)$ and $\varphi_2(\mathbf{x}_2)$ versus observed predictors displayed in the left middle panels. Linearization of partial residuals shown in far right panels. Transformations displayed by blue curves, and independently fitted smoothing splines by red curves.*

## 3. COMPUTATION OF REGRESSION WEIGHTS AND TRANSFORMATION PARAMETERS

Since the predictor variables are usually correlated in the regression problem (2), the optimal transformations $\varphi_k(\mathbf{x}_k)$ in (4) and quantifications $\mathbf{v}_k$ in (5) are also interdependent. For the moment, we assume the transformation of the outcome $\vartheta(\mathbf{y})$ to be fixed. To solve for each $\varphi_k(\mathbf{x}_k)$, we separate a variable and its weight from the linear combination of predictors, isolating the current target part $\beta_k \varphi_k(\mathbf{x}_k)$ from the remainder, denoted as $\sum_{l \neq k} \beta_l \varphi_l(\mathbf{x}_l)$. As mentioned in the introduction, this approach has been called *block modeling*, the Gauss-Seidel algorithm, *alternating least squares*, *backfitting* and most recently as *coordinate descent*, and goes as follows. We rewrite the loss function as

$$L(\beta, \varphi) = ||\vartheta(\mathbf{y}) - \sum_{l \neq k} \beta_l \varphi_l(\mathbf{x}_l) - \beta_k \varphi_k(\mathbf{x}_k)||^2, \qquad (9)$$

where $\beta$ again denotes $\{\beta_k\}\}_{k=1}^{P}$ and $\varphi$ stands for $\{\varphi_k(\mathbf{x}_k)\}_{k=1}^{P}$. Then we turn the original multivariate problem into a univariate one; first we define an auxiliary variable $\mathbf{u}_k$:

$$\mathbf{u}_k = \vartheta(\mathbf{y}) - \sum_{l \neq k} \beta_l \varphi_l(\mathbf{x}_l), \tag{10}$$

thus $\mathbf{u}_k$ is the partial residual. Next we simply have to minimize

$$L(\beta_k, \varphi_k) = ||\mathbf{u}_k - \beta_k \varphi_k(\mathbf{x}_k)||^2, \tag{11}$$

which is a function of $\beta_k$ and $\varphi_k(\mathbf{x}_k)$ only. The standardization of the transformed variable $\varphi_k(\mathbf{x}_k)$ allows us to compute the regression weight $\beta_k$ separately from the transformation. The current value for the regression weight $\beta_k$ is obtained as

$$\tilde{\beta}_k = \mathbf{u}'_k \varphi_k(\mathbf{x}_k). \tag{12}$$

Next we minimize (11) over all $\varphi_k(\mathbf{x}_k) \in \mathbb{C}_k(\mathbf{x}_k)$, where $\mathbb{C}_k(\mathbf{x}_k)$ specifies the cone that contains all admissible transformations of the variable $X_k$. In the case of a nominal transformation, the cone $\mathbb{C}_k(\mathbf{x}_k)$ is defined by

$$\mathbb{C}_k(\mathbf{x}_k) \equiv \{\varphi_k(\mathbf{x}_k)|\varphi_k(\mathbf{x}_k) = \mathbf{G}_k \mathbf{v}_k\}, \tag{13}$$

and we define the metric projection $P_{\mathbb{C}_k(\mathbf{x}_k)}$ as

$$P_{\mathbb{C}_k(\mathbf{x}_k)} \equiv \min_{\mathbf{v}_k} ||\mathbf{u}_k - \beta_k \mathbf{G}_k \mathbf{v}_k||^2. \tag{14}$$

This metric projection ensures that objects in the same category according to variable $k$ obtain the same quantification in the transformed variable $\varphi_k(\mathbf{x}_k) = \mathbf{G}_k \mathbf{v}_k$, and this is ensured by setting

$$\tilde{\mathbf{v}}_k = \beta_k^{-1} \mathbf{D}_k^{-1} \mathbf{G}'_k \mathbf{u}_k, \tag{15}$$

where $\mathbf{D}_k = \mathbf{G}'_k \mathbf{G}_k$, a diagonal matrix with the marginal frequencies of the categories on the main diagonal. Actually, only the sign of $\beta_k$ is needed because the transformed variable $\varphi_k(\mathbf{x}_k)$ will be standardized. The latter is ensured by setting

$$\hat{\mathbf{v}}_k = N^{1/2} \tilde{\mathbf{v}}_k (\tilde{\mathbf{v}}'_k \mathbf{D}_k \tilde{\mathbf{v}}_k)^{-1/2}. \tag{16}$$

For ordinal transformations, the cone $\mathbb{C}_k$ that contains all monotonic transformations of $X_k$ is defined by

$$\mathbb{C}_k(\mathbf{x}_k) \equiv \{\varphi_k(\mathbf{x}_k)|\varphi_k(\mathbf{x}_k) = mon(\mathbf{x}_k)\}, \tag{17}$$

where $mon(\mathbf{x}_k)$ denotes a least squares monotonic transformation of $X_k$. The metric projection is written as

$$P_{\mathbb{C}_k(\mathbf{x}_k)} \equiv \min_{mon(\mathbf{x}_k)} ||\mathbf{u}_k - \beta_k mon(\mathbf{x}_k)||^2, \tag{18}$$

which amounts to applying *monotonic (isotonic) regression* of $sign(\beta_k^{-1})\mathbf{u}_k$ onto $\mathbf{x}_k$, written as $mon(sign(\beta_k^{-1})\mathbf{u}_k, \mathbf{x}_k)$, and standardizing the result. The monotonic regression can either be increasing or decreasing, whichever gives the smaller loss value; if applicable, the sign of $\beta_k$ has to be adjusted.

In the case of spline transformations, the cone is defined by

$$\mathbb{C}_k(\mathbf{x}_k) \equiv \{\varphi_k(\mathbf{x}_k)|\varphi_k(\mathbf{x}_k) = splin(\mathbf{x}_k)\}, \tag{19}$$

where the metric projection is written as

$$P_{\mathbb{C}_k(\mathbf{x}_k)} \equiv \min_{splin(\mathbf{x}_k)} ||\mathbf{u}_k - \beta_k splin(\mathbf{x}_k)||^2. \tag{20}$$

The term $splin(\mathbf{x}_k)$ denotes a smooth transformation of the predictor $X_k$ using splines. One possibility is to construct an $I$-spline basis matrix $S_k(\mathbf{x}_k)$ (see Ramsay (1988) for details), and having $\mathbf{S}_k = S_k(\mathbf{x}_k)$, we minimize

$$L(\mathbf{b}_k) = ||\mathbf{u}_k - \beta_k \mathbf{S}_k \mathbf{b}_k||^2, \tag{21}$$

over $\mathbf{b}_k = \{b_t^k\}_{t=1}^{T_k}$, the $T_k$-vector with spline coefficients that have to be estimated, and where $T_k$ is dependent on the degree of the spline and the number of interior knots. If the $I$-spline transformation does not have to follow the order of the values in $X_k$, we can compute the analytical solution for $\mathbf{b}_k$ directly, since (21) is a straightforward regression problem, with the columns of $\mathbf{S}_k = \mathbf{s}_{t=1}^{T_k}$ as independent variables. If, however, the $I$-spline transformation is required to be monotonic with $\mathbf{x}_k$, we have to minimize (21) under the restriction that the vector $\mathbf{b}_k$ with spline coefficients contains nonnegative elements. This constrained optimization problem can be solved by applying the one-variable-at-a-time strategy here as well. Thus, the problem is further partitioned by isolating the $t$th column of the spline basis matrix $\mathbf{S}_k$ (denoted by $\mathbf{s}_t^k$) and the $t$th element ($b_t^k$) of the spline coefficient vector $\mathbf{b}_k$ from the remaining elements $\{b_r^k\}_{r \neq t}$. Next, we minimize iteratively

$$L(b_t^k) = ||(\mathbf{u}_k - \beta_k \sum_{r \neq t} b_r^k \mathbf{s}_r^k) - \beta_k b_t^k \mathbf{s}_t^k||^2 \tag{22}$$

over $b_t^k \geq 0$, for $t = 1, ..., T_k$. (There is a complication if we take the normalization condition $\mathbf{b}_k' \mathbf{S}_k' \mathbf{S}_k \mathbf{b}_k = N$ into account that ensures that the transformed variable is standardized; how this problem is solved can be found in Groenen, Van Os, and J.J. (2000).) For completeness, we mention the linear transformation, which defines the cone as

$$\mathbb{C}_k(\mathbf{x}_k) \equiv \{\varphi_k(\mathbf{x}_k)|\varphi_k(\mathbf{x}_k) = stand(\mathbf{x}_k)\}, \tag{23}$$

which amounts to using a standardized version of $\mathbf{x}_k$. When updates for both $\beta_k$ and $\varphi_k(\mathbf{x}_k)$ have been found, we estimate both the transformation and associated regression coefficient for each of the other predictors, one at-a-time.

When all coefficients and variable transformations have been updated in this way, we could transform the outcome variable as well, for which we have a similar set of transformation options available as for the predictor variables. Writing the loss function as a function of the outcome variable only, amounts to:

$$L(\vartheta) = ||\vartheta(\mathbf{y}) - \sum_{k=1}^{P} \beta_k \varphi_k(\mathbf{x}_k)||^2 = ||\vartheta(\mathbf{y}) - \mathbf{z}||^2, \tag{24}$$

where $\vartheta(\mathbf{y})$ denotes a transformation of the response variable $Y$. If a spline transformation is chosen, the transformation can easily be found by the metric projection

$$P_{\mathbb{C}(\mathbf{y})} \equiv \min_{splin(\mathbf{y})} ||\mathbf{z} - splin(\mathbf{y})||^2.$$

Although the full set of transformations is available for the outcome variable, if the outcome is continuous, we choose in practice a linear transformation, or a monotonic (spline) transformation that uses only very few degrees of freedom. If the outcome is ordered categorical, we choose a monotonic step function. This strategy is chosen for reasons of interpretability of the final regression model. A possible nonlinear relation between (a linear combination of) predictor variables and the outcome is preferably taken care of by nonlinear transformation of the predictors. An exception would be made for the case when the outcome variable is unordered categorical, because in that case optimal scaling is equivalent to classical linear discriminant analysis.

## 4. APPLICATIONS TO EMPIRICAL DATA

4.1 Regression with mixed measurement level predictors

The data used in this example were collected at the Leiden Cytology and Pathology Laboratory, and concern characteristics of cells obtained from patients with various grades of cervical preoplasia and neoplasia. To obtain the samples, taken from the ectocervix as well as the endocervix, special sampling and preparation techniques were used. The correct histological diagnosis was known by a subsequently taken biopsy. A subset of the data has been previously analysed in Meulman, Zeppa, Boon, and Rietveld (1992) and Friedman and Meulman (2003), and contains, according to the histological diagnosis, 50 cases with mild dysplasia (histological group 1), 50 cases with moderate dysplasia (histological group 2), 50 cases with severe dysplasia (histological group 3), and 50 cases with carcinoma in situ (histological group 4). The number of cases with invasive squamous cell carcinoma (histological group 5) is 42. For each of the 242 cases, seven qualitative features of the cells were determined. The features were rated by a pathologist on a scale ranging from 1 (normal) to 4 (very abnormal); so these seven variables are ordered categorical. The features under consideration are *Nuclear Shape, Nuclear Irregularity, Chromatin Pattern, Chromatin Distribution, Nucleolar Irregularity, Nucleus/Nucleolus Ratio*, and *Nucleus/Cytoplasm Ratio*. In addition, four quantitative features of each sample were established: *Number of Abnormal Cells per Fragment* (mean values), *Total Number of Abnormal Cells, Number of Mitoses*, and *Number of Nucleoli* (mean values).

From the earlier analyses mentioned above, it is known that this data set is noisy, and accurate prediction of the outcome is thereby difficult.

In a first comparison, we fit three different sets of transformations. In the first set (models 1 to 3), transformations of the response and the quantitative predictors are linear; the qualitative predictors obtain a linear, nominal, and ordinal transformation, respectively. Because the results indicate that ordinal transformations for the qualitative variables are most appropriate, we also fit those in the second set of models. In addition, we fit nonlinear spline functions for the quantitative variables, both nonmonotonic and monotonic, and varying the number of interior knots. The results show that monotone functions are preferred over nonmonotone functions. In the third set, we apply the same transformations, but now also an ordinal transformation of the outcome variable (diagnosis), and obtain even better results for the prediction accuracy, both in the cross-validation as for the test data. Increasing the number of knots for the monotonic splines is hardly worthwhile.

*Table 2. Prediction error for different sets of transformations. Mean values over 12 subsets of the data.*

|   | Predictors | | Response | Mean prediction error(sd) | | |
|---|---|---|---|---|---|---|
|   | Qual | Quant |   | training data | cross-validation | test data |
| 1 | linear | linear | linear | .254 | .288(.026) | .279(.076) |
| 2 | nominal | linear | linear | .216 | .274(.027) | .269(.080) |
| 3 | ordinal | linear | linear | .217 | .270(.027) | .266(.079) |
| 4 | ordinal | spline(nmon,2,2) | linear | .148 | .203(.018) | .195(.055) |
| 5 | ordinal | spline(nmon,2,1) | linear | .150 | .197(.017) | .189(.052) |
| 6 | ordinal | spline(mono,2,1) | linear | .153 | .191(.017) | .183(.051) |
| 7 | ordinal | spline(mono,2,2) | linear | .150 | .189(.017) | .182(.051) |
| 8 | ordinal | spline(nmon,2,2) | ordinal | .125 | .179(.021) | .179(.060) |
| 9 | ordinal | spline(nmon,2,1) | ordinal | .128 | .175(.020) | .174(.059) |
| 10 | ordinal | spline(mono,2,1) | ordinal | .134 | .175(.019) | .174(.055) |
| 11 | ordinal | spline(mono,2,3) | ordinal | .127 | .171(.019) | .170(.058) |
| 12 | ordinal | spline(mono,2,2) | ordinal | .128 | .169(.019) | .167(.056) |

First, in the training phase, a number of different combinations of transformation have been compared, starting with linear transformations for all predictors (most restricted) to nonlinear transformations. All models were cross-validated. The total number of objects is 242, and 20 objects were set apart in each step of the process. From the remaining 222 objects, 202 objects at a time were used as training data, and 11-fold cross-validation (leave-out 20) was applied to obtain the expected prediction error for the training data. This procedure was repeated for each of the 12 sets of randomly selected splits of objects held apart to obtain the prediction error for the test data. In the test phase, we fix the shape of the transformation, but not its parameters. Thus the parameters are estimated again in each step. We leave out 20 of the 242 cases in each of the 12 11-fold cross-validations in the test phase. Then, to test the results, the transformations from each of the training samples (of size 202) were used on the test set. The mean prediction error and the standard deviations are given in Table 1.

The cross-validation results show that a model with all transformations linear was least successful; better results were obtained when nonlinear transformations were applied, first for the qualitative predictors, allowing for ordinal transformation), and next also for the quantitative predictors. For the latter, models that were fitted included both nonmonotonic and monotonic cubic splines, with either one or two interior knots. From these results a particular selection is depicted in Figure 1. The first seven boxplots show the apparent prediction errror for the training data, and it is clear that the differences between linear transformations for the quantitative predictors (tr1, tr2, tr3) and nonlinear transformations (tr4, tr7, tr8, tr12) are large, and those between nonmonotonic and monotonic transformations (tr4 and tr7, and tr8 and tr12) are small. There is a difference, however, between linear and ordinal transformation of the (categorical) outcome (tr 4 and tr7 versus tr8 and tr12). The next seven boxplots give the cross-validated prediction error, and these show that all predictors should be transformed with monotonic functions (cv4 and cv8, versus cv7 and cv12). The last six boxplots show the prediction error for the test data, which were not used in any of the other analyses. (Note that we have omitted model 2 from the comparison, because its results can hardly be distinguished from model 3.) The variation is obviously much larger, as was also the case for the standard deviations given in Table 1. The median prediction error is, however, completely comparable to the one obtained by the cross-validation, and the overall pattern shows again that monotonic transformations should be preferred throughout.

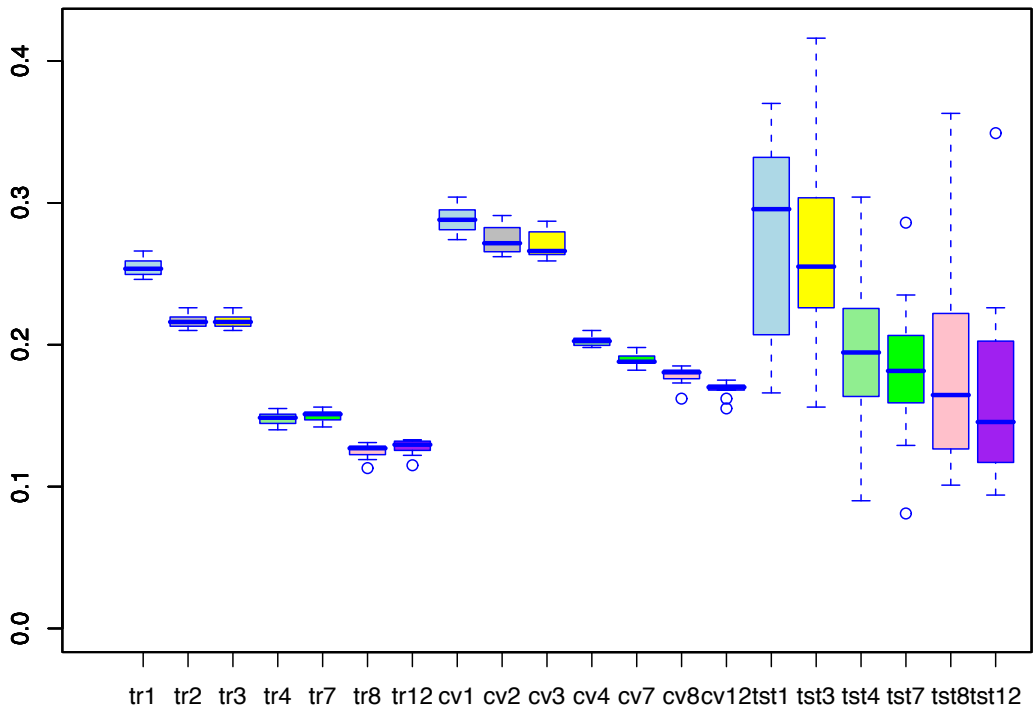

*Figure 2. Comparing different models for the Cervix data with respect to mean prediction error obtained for traing data (tr), cross-validation (cv), and test data (tst); labels refer to the models in Table 1.*

In this paragraph, we give diagnostics for predictor correlation matrices for a hierarchy of models with increasing number of degrees of freedom due to different sets of transformation. The sets are 1: All predictors and outcome linear. 2: Outcome linear, categorical predictors ordinal, quantitative predictors monotonic spline (2,2). 3: Like 2, but quantitative predictors nonmonotonic spline (2,2). 4: Like 3, but with quantitative predictors nonmonotonic spline (3,3). 5: Like 4, but quantitative predictors nominal transformation. 6: Like 5, but outcome ordinal. In Figure 3, we display the smallest eigenvalues (left panel), and the corresponding log-determinant divergence from independence (middle panel). The values for both diagnostics differ considerably among the six models, where the size differences between model 1 and model 6 are (almost) of the order 3 (.136 and .393, respectively, for SMEV, and 4.52 and 1.81, respectively, for DLD). In the panel at the right, we display the average dependence for the predictors, which can easily be computed from the inverse of the eigenvalues of the predictor correlation matrix. If we write $\mathbf{R} = \mathbf{L}\mathbf{\Lambda}\mathbf{L}'$ then $\mathbf{R^{-1}} = \mathbf{L}\mathbf{\Lambda^{-1}}\mathbf{L}'$, and $trace(\mathbf{R^{-1}}) = trace(\mathbf{\Lambda^{-1}})$. (The diagonal of $\mathbf{\Lambda^{-1}}$ contains the eigenvalues of $\mathbf{R^{-1}}$ in reversed order.) The three diagnostics show overall the same pattern for the different models. The smallest eigenvalues (left panel) increase in each step, and divergence from independence (middle panel) and average predictor dependency (right panel) decrease. The largest step is taken when going from the first to the second model in the hierarchy (which are models 1 (predictors linear) and 7 (optimally scaled predictors) in Table 2, respectively). The smallest eigenvalue plot shows a substantial increase between 5 and 6, which models are identical except for the fact that in the sixth model the outcome is ordinal instead of linear. The divergence from independence and average predictor dependency show a drop when the outcome is transformed.

To conclude this example, we display the optimal quantifications for Model 12 in Table 2 in the transformation plots in Figure 4. The black circles represent the category quantifications from the analyses for the 12 training data sets, each consisting of 222 objects. The red lines connects the average of the quantifications in the 12 training sets. We observe the following. With respect to the transformation of Diagnosis, the biggest step is between the categories 3 and 4. This has a very clear clinical counterpart, since it is the difference between severe dysplasia and the first class of cancer (carcinoma in situ). Apparently, this departure from linearity has a positive effect on the prediction accuracy. Steps have about equal size for Nucl_Shape, and Chrom_Pat, but not for the five other qualitative predictors. Transformations for #Abn_Cells, Tot#_Abn, #Mitoses are smooth, but the one for #Nucleoli is not. Overall, the quantifications for the 12 training data sets are remarkable stable.

### 4.2 OS Regression with several sets of transformations for car data

In the second application of OS regression to real data, we analyze data on 405 cars, where seven predictors give various properties of the cars, aimed to predict the outcome "Time to accelerate from 0 to 60 mph" (in seconds). (The data were taken from the SPSS data library.) The seven predictors are 1. Miles per gallon, 2.Engine displacement (in cubic inches), the volume of the engine cylinder, 3. Horsepower, 4. Weights (in lbs), 5. Number of cylinders, 6. Year of the particular model (modulo 100), and 7. Country of origin (US, Europe, or Japan). We discuss four analyses. In the first analysis, the categorical predictors "Number of cylinders", "Model year", and "Country of Origin" are treated nominally, all other predictors numerically. After inspection of the partial residual plots, in the second analysis, "Miles per gallon" was subsequently transformed by a nonmonotonic spline (cubic, 3 internal knots), and the other quantitative predictors by monotonic splines (3,3).

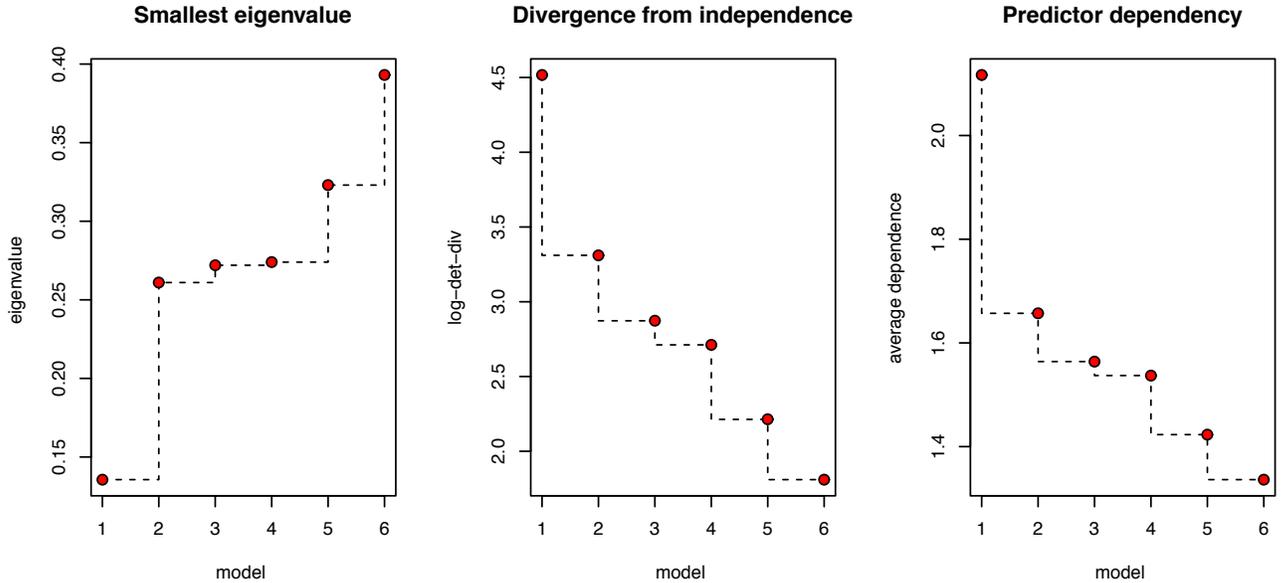

*Figure 3. Smallest eigenvalues of six predictor correlation matrices (left panel), difference between $\mathbf{R}$ and $\mathbf{I}$ by log determinants (middle panel), and average dependence, the average multiple correlation between each predictor and the other predictors.*

### 4.2.1 Transforming the outcome variable

On the basis of the scatter of residual points in the plot of $\hat{\mathbf{y}}$ versus $\mathbf{y}$, showing overall nonlinearity (Figure 5, left panels), it was decided in the 3rd analysis to fit a cubic monotonic spline (3,0) $\vartheta(\mathbf{y})$ to $\mathbf{y}$ (upper middle panel). The resulting scatter of residual points (upper right panel) becomes linear for the higher values for $\vartheta(\mathbf{y})$, but for the small values some nonlinearity remains. After fitting a cubic nonmonotonic spline with 3 internal knots (4th analysis, bottom middle panel), this remaining nonlinearity disappears (bottom right panel).

The main question that remains is, of course, whether the nonmonotonic transformation of the outcome variable is spurious; i.e. that by overfitting the expected prediction error (EPE) will increase. In Table 3 we see that this is not the case. By default, the $r^2$ increases with each new analysis since subsequent analyses are less restricted. Surprisingly, however, we note that model 3, with nonmonotonic transformation of the outcome, has smaller expected prediction error and also smaller standard deviation.

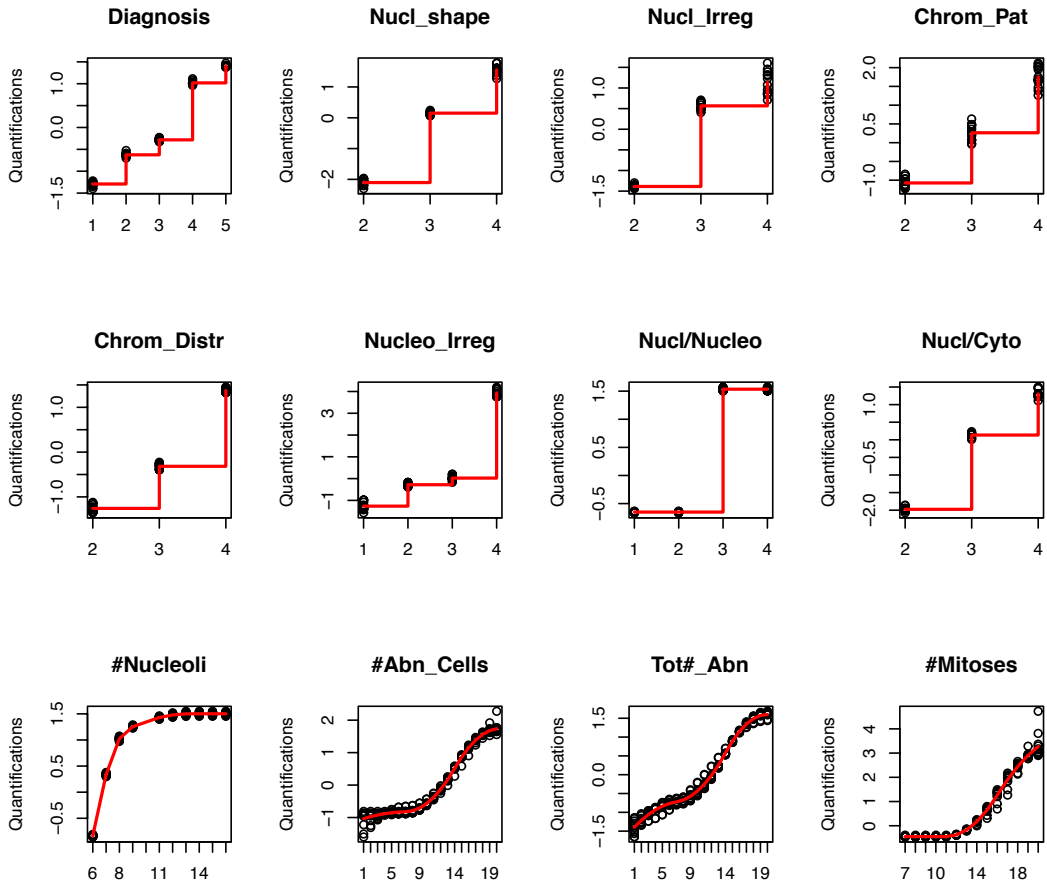

*Figure 4. Ordinal transformations for diagnosis (response, five categories), and seven categorical predictors. Monotonic quadratic spline transformations with two interior knots for the four quantitative predictors.*

Figure 6 shows the transformations for *mpg, engine, horse power* and *weight*. The red curve represents the transformations for analysis 4, and the blue curve the transformation for analysis 3. It is clear that the transformations in the different analyses are remarkably similar, although the transformations for the outcome are very different.

*Table 3. Fit and expected prediction error for Car data*

*1: numeric predictors linear, categorical predictors nominal*

*2: mpg nonmonotonic, other numeric predictors monotonic splines*

*3. like 2, and monotonic spline outcome*

*3: like 2, nonmonotonic spline outcome.*

| Outcome | $r^2$ | EPE(s.e.) | SMEV | DLD | PRED |
|---|---|---|---|---|---|
| 1. Linear | .641 | .403(.037) | .052 | 7.055 | 5.691 |
| 2. Linear | .757 | .309(.028) | .058 | 4.676 | 4.334 |
| 3. Monotonic | .796 | .261(.036) | .080 | 4.754 | 3.925 |
| 4. Nonmonotonic | .833 | .216(.027) | .097 | 4.559 | 3.596 |

*Table 4. Regression coefficients for the car data*

*Model 1: quantitative predictors numerical, qualitative predictors nominal*

*Model 2: mpg nonmonotonic, other quantitative predictors monotonic splines*

*Model 3: nonmonotonic transformation of outcome Acceleration.*

| X | $1:\beta$(s.e.) | $2:\beta$(s.e.) | $3:\beta$(s.e.) | 1: F | 2: F | 3: F | 1: Tol | 2: Tol | 3: Tol |
|---|---|---|---|---|---|---|---|---|---|
| 1 | -0.006(.095) | 0.229(.072) | 0.242(.072) | .004 | 10.180 | 11.273 | .290 | ..646 | .644 |
| 2 | -0.201(.216) | -0.667(.229) | -0.650(.164) | .866 | 8.458 | 15.600 | .074 | .103 | .197 |
| 3 | -1.186(.094) | -1.197(.095) | -0.974(.141) | 159.181 | 157.349 | 47.852 | .153 | .188 | .135 |
| 4 | 0.919(.167) | 1.253(.146) | 0.984(.168) | 30.352 | 73.392 | 34.227 | .108 | .106 | .158 |
| 5 | 0.199(.088) | 0.236(.103) | 0.242(.081) | 5.067 | 5.311 | 9.026 | .265 | .496 | .405 |
| 6 | 0.098(.029) | 0.103(.028) | 0.143(.034) | 11.838 | 13.781 | 17.525 | .784 | .937 | .815 |
| 7 | 0.022(.033) | 0.075(.037) | 0.040(.023) | .450 | 4.128 | 3.060 | .504 | .838 | .928 |

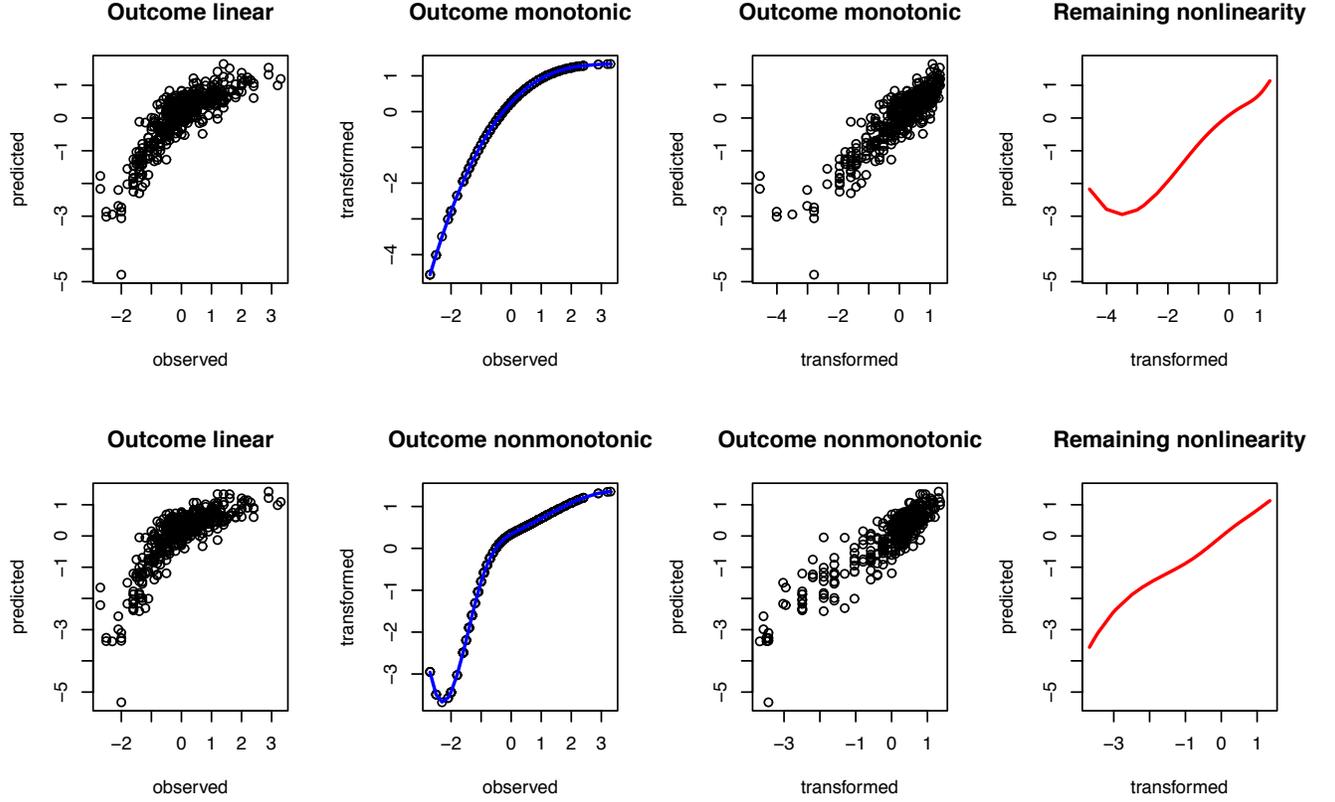

*Figure 5. Transformation and prediction of outcome in Car Data.*

## 5. OPTIMAL SCALING REGRESSION WITH LASSO, RIDGE, AND ELASTIC NET PENALTIES

Ridge regression, the Lasso, and the Elastic Net constrain the size of the regression coefficients by setting a maximum on the sum of the squared coefficients (Ridge), or the sum of absolute values of the coefficients (Lasso), or on both these sums (Elastic Net). The loss functions are constrained versions of the ordinary least squares (OLS) regression loss function, and are written as follows below. For Ridge, we write

$$L^{\text{ridge}} = \|\mathbf{y} - \sum_{k=1}^{P} b_k \mathbf{x}_k\|^2, \text{ subject to } \sum_{k=1}^{P} b_k^2 \leq t_2, \tag{25}$$

with $t_2$ a tuning parameter with respect to the sum of squares of the $b_k$, and its value has to be determined in the optimization process. The Lasso constrains the sum of the absolute values of the regression coefficients:

$$L^{\text{lasso}} = \|\mathbf{y} - \sum_{k=1}^{P} b_k \mathbf{x}_k\|^2, \text{ subject to } \sum_{k=1}^{P} |b_k| \leq t_1. \tag{26}$$

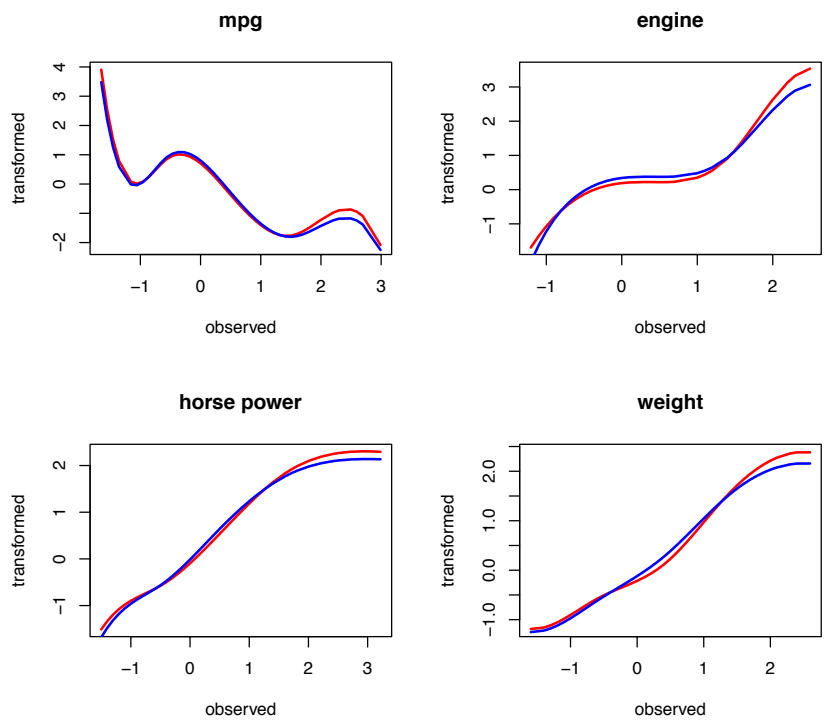

*Figure 6. Two sets of transformations of quantitative predictors in Car data. Blue: with outcome monotonic. Red: with outcome nonmonotonic.*

The elastic net combines the ridge and Lasso contraints:

$$L^{\text{enet}} = \|\mathbf{y} - \sum_{k=1}^{P} b_k \mathbf{x}_k\|^2, \text{ subject to } \sum_{k=1}^{P} b_k^2 \leq t_2 \text{ and } \sum_{k=1}^{P} |b_k| \leq t_1. \quad (27)$$

Applying such a constraint is equivalent to penalizing the sum of squares or the sum of absolute values of the coefficients, because there is a one-to-one relation between the value of the maximum of the sum and the penalty value. Thus, the Ridge, Lasso, and Elastic Net loss functions can also be written as penalized ordinary least squares (OLS) regression loss functions, and amount to:

$$L^{\text{ridge}}(\beta) = \|\mathbf{y} - \sum_{k=1}^{P} \beta_k \mathbf{x}_k\|^2 + \lambda_2 \sum_{k=1}^{P} \beta_k^2, \quad (28)$$

$$L^{\text{lasso}}(\beta) = \|\mathbf{y} - \sum_{k=1}^{P} \beta_k \mathbf{x}_k)\|^2 + \lambda_1 \sum_{k=1}^{P} \text{sign}(\beta_k)\beta_k, \quad (29)$$

$$L^{\text{e-net}}(\beta) = \|\mathbf{y} - \sum_{k=1}^{P} \beta_k \mathbf{x}_k\|^2 + \lambda_1 \sum_{k=1}^{P} \text{sign}(\beta_k)\beta_k + \lambda_2 \sum_{k=1}^{P} \beta_k^2 \quad (30)$$

Here $\lambda_1$ denotes the strength of the Lasso penalty, and $\lambda_2$ of the Ridge penalty. Minimization of the Ridge loss function (28) has an analytic solution:

$$\hat{\beta}^{\text{ridge}} = (\mathbf{X}'\mathbf{X} + \lambda_2 \mathbf{I})^{-1}\mathbf{X}'\mathbf{y}. \quad (31)$$

The estimates of the regression coefficients for the Lasso are

$$\hat{\beta}^{\text{lasso}} = (\mathbf{X}'\mathbf{X})^{-1}(\mathbf{X}'\mathbf{y} - \frac{\lambda_1}{2}\mathbf{w}), \quad (32)$$

where the elements $w_k$ of $\mathbf{w}$ are either $+1$ or $-1$, depending on the sign of the corresponding regression coefficient $\beta_k$. Obtaining the Lasso coefficients is a least squares problem with $2^P$ inequality constraints (there are $2^P$ possible sign patterns for the coefficients), and was efficiently solved for by the LARS algorithm. For the Elastic Net, the regression coefficients are estimated as

$$\hat{\beta}^{\text{e-net}} = (\mathbf{X}'\mathbf{X} + \lambda_2 \mathbf{I})^{-1}(\mathbf{X}'\mathbf{y} - \frac{\lambda_1}{2}\mathbf{w}),$$

and minimization of this loss function is much like minimizing the Lasso loss function, and the entire Elastic Net regularization paths can be estimated almost as efficiently as the Lasso paths with the LARS-ENet algorithm (Zou and Hastie 2005).

When the predictors are orthogonal, the Lasso estimates have a simple form:

$$\beta_k^{\text{lasso}} = (\hat{\beta}_k - \frac{\lambda_1}{2} w_k)_+, \quad (33)$$

where $w_k$ is $+1$ or $-1$ depending on the sign of the corresponding $\beta_k$, and $(\cdot)_+$ denotes truncation at zero.

## 5.1 Obtaining regularized regression coefficients in Optimal Scaling regression

As was shown in Section 2, the OS algorithm estimates the transformations and regression coefficients one at a time, and it removes the effect of the other predictors from the outcome when estimating the coefficient for a particular predictor by using (10), (11), and (12). To include regularization penalties, the same strategy works here as well. The three loss functions for regularized Optimal Scaling (with Ridge, the Lasso, and the Elastic Net, respectively) are written as

$$L^{\text{ridge}}(\beta) = \|\mathbf{y} - \sum_{l \neq k} \beta_l \varphi_l(\mathbf{x}_l) - \beta_k \varphi_k(\mathbf{x}_k))\|^2 + \lambda_2 \sum_{l \neq k} \beta_l^2 + \lambda_2 \beta_k^2$$

$$L^{\text{lasso}}(\beta) = \|\mathbf{y} - \sum_{l \neq k} \beta_l \varphi_l(\mathbf{x}_l) - \beta_k \varphi_k(\mathbf{x}_k))\|^2 + \lambda_1 \sum_{l \neq k} |\beta_l| + \lambda_1 \text{sign}(\beta_k)\beta_k$$

$$L^{\text{e-net}}(\beta) = \|\mathbf{y} - \sum_{l \neq k} \beta_l \varphi_l(\mathbf{x}_l) - \beta_k \varphi_k(\mathbf{x}_k))\|^2 + \lambda_2 \sum_{l \neq k} \beta_l^2 + \lambda_1 \sum_{l \neq k} |\beta_l|$$
$$+ \lambda_2 \beta_k^2 + \lambda_1 \text{sign}(\beta_k)\beta_k.$$

The crucial result of this approach is that the estimates of the regularized coefficients $\beta_k^{\text{ridge}}$ and/or $\beta_k^{\text{lasso}}$ can be computed as if the predictors were uncorrelated. So, incorporating regularization in the OS regression loss function only requires adjusting the estimation of the regression coefficients for the Lasso as in (33), and this amounts to

$$\hat{\beta}_k^{\text{lasso}} = \begin{cases} \tilde{\beta}_k - \frac{\lambda_1}{2} & \text{if } \tilde{\beta}_k > 0 \text{ and } \frac{\lambda_1}{2} < \tilde{\beta}_k \\ \tilde{\beta}_k + \frac{\lambda_1}{2} & \text{if } \tilde{\beta}_k < 0 \text{ and } \frac{\lambda_1}{2} < |\tilde{\beta}_k| \\ 0 & \text{otherwise} \end{cases} \quad (34)$$

for the Lasso. Since the estimate for the Ridge coefficeient is given by

$$\hat{\beta}_k^{\text{ridge}} = \tilde{\beta}_k/(1 + \lambda_2), \quad (35)$$

it turns out the the coefficients in Elastic Net regularization are found as

$$\hat{\beta}_k^{*\text{e-net}} = \begin{cases} \frac{\beta_k - \frac{\lambda_1}{2}}{1+\lambda_2} & \text{if } \tilde{\beta}_k > 0 \text{ and } \frac{\lambda_1}{2} < \tilde{\beta}_k \\ \frac{\beta_k + \frac{\lambda_1}{2}}{1+\lambda_2} & \text{if } \tilde{\beta}_k < 0 \text{ and } \frac{\lambda_1}{2} < |\tilde{\beta}_k| \\ 0 & \text{otherwise .} \end{cases} \quad (36)$$

Here $\tilde{\beta}_k$ is the simple update $\tilde{\beta}_k = \mathbf{u}_k' \varphi_k(\mathbf{x}_k)$ from (12). We correct for the double amount of shrinkage in the estimation of the Elastic Net coefficients by rescaling the coefficients $\hat{\beta}_k^{*\text{e-net}}$ after convergence:

$$\hat{\beta}_k^{\text{e-net}} = \hat{\beta}_k^{*\text{e-net}}(1 + \lambda_2), \quad (37)$$

as suggested by Zou and Hastie (2005).

### 5.2 Selection of the optimal value of the penalty parameter(s)

For selecting the optimal value of the penalty parameter(s), the Expected Prediction Error (EPE) for each (combination of) penalty value(s) has to be estimated. To estimate the EPE, analytic methods like Generalized Cross Validation (GCV; Golub, Heath, and Wahba 1979), AIC, or BIC can be used or a resampling method, such as cross validation or bootstrapping. Using a resampling method for model selection is time consuming, because it has to be repeated for each (combination of) penalty value(s). However, resampling has two major advantages over analytic methods: it does not require estimation of the degrees of freedom involved, and it also works when $P > N$. Application of the .632 bootstrap or 10-fold cross-validation can be made much less time-consuming by not assessing the expected prediction error for all values of the penalty parameter. We have observed that a plot of the estimates of the expected prediction error as a function of the model complexity usually shows a regular curve: we obtain the highest error estimates for the highest values of the penalty parameter, and the values of the error estimates decrease with decreasing values of the penalty parameter, until we reach the minimum. From that point, the error estimates increase again until we reach the point for the zero penalty term. Thus, the model selection procedure can be made much more efficient by doing the full analysis in two phases. In the first phase, the region of the optimal values on the path is determined by using a rather big step size for consecutive values of the penalty parameter. In the second phase, the search is limited to this region (that contains the minimum), and the optimal values themselves are determined by taking much smaller steps. Model selection involves resampling in combination with the one-standard-error rule: the most parsimonious model within one standard error of the minimum is selected. In our applications we use both cross-validation and the .632 bootstrap method (Efron 1983) that theoretically gives a better estimate of the Expected Prediction Error than the standard bootstrap. The details of how to use the .632 bootstrap with the CATREG program are extensively described in Van der Kooij (2007).

### 5.3 Regularization for Data Example 1

We apply the three forms of regularization to the small data example 1, and from those we choose the model that has the smallest expected prediction error within 1 standard deviation from the optimal model. For the analysis with linear transformations, this turns out to be the Lasso regularization, where the lasso penalty is 0.80, and where the second predictor variable is left out of the analysis. If we include transformation of the second predictor only (since it was omitted from the first analysis), the Ridge regularization is selected, with a penalty of 1.10, and both predictors in the model, with regression coefficients .255 and .237, respectively. Regression coefficients are very similar, the Expected Prediction Error decreases, and so is the correlation between the two predictors. The tolerance increases. Next, if we allow both predictors to be transformed, the first predictor becomes dominant again, the Expected Prediction Error becomes very small, as well as its standard deviation, while the dependence between transformed predictors and the tolerance are comparable to the previous analysis. However, the ridge penalty in the chosen model is merely 0.10, thus results are very similar to those of the OS analysis without regularization in Table 1.

*Table 5. Data Example 1: Best regularized model for three combinations of transformations.*

| Transformation | $r^2$ | $\beta_1$(s.e.) | $\beta_2$(s.e.) | EPE(s.e.) | $r(x_1, x_2)$ | Tol | ridge | lasso |
|---|---|---|---|---|---|---|---|---|
| 1. $\lin(x_1)$, $\lin(x_2)$ | .379 | .215(.022) | —(—) | .826(.232) | .706 | .502 | 0.00 | 0.80 |
| 2. $\lin(x_1)$, $\splin(x_2)$ | .538 | .255(.009) | .237(.014) | .607(.199) | .340 | .885 | 1.10 | 0.00 |
| 3. $\spl(x_1)$, $\spl(x_2)$ | .853 | .746(.029) | .225(.024) | .152(.007) | .354 | .874 | 0.10 | 0.00 |

*Table 6. Social indicator variables for the United States.*

| Label | Description |
|---|---|
| POPUL | 1975 population in thousands |
| INCOME | Per capita income in dollars |
| ILLIT | Illiteracy rate in percent of population |
| LIFE | Life expectancy in years |
| HOMIC | 1976 homicide and non-negligent manslaughter (per 1000) |
| SCHOOL | Percent of population over age 25 who are high school graduates |
| FREEZE | Average number of days of the year with temperatures below zero |

## 6. APPLICATION OF REGULARIZATION AND OPTIMAL SCALING TO THE UNITED STATES DATA

The States Data example concerns data analyzed in Meulman (1986) with the predictor variables taken from Wainer and Thissen (1981) who used seven social indicator statistics in order to re-examine the Angoff and Mencken (1931) search for "The Worst American State". The outcome variable gives the percentage of failure on a nation-wide test. The description of the variables is given in Table 7.

To combine transformation with estimation of the expected prediction error using the .632 bootstrap, the 50 values in the original variables were binned into 15 categories, following a uniform distribution as closely as possible. It was already shown in Meulman (1986) that the original data contain some serious nonlinearities, e.g. the relation between POPUL on the one hand, and INCOME and ILLIT on the other hand.

The first model option is the base analysis, since it uses neither optimal scaling nor regularization, and the expected prediction error (estimated with 50 samples for the .632 bootstrap) is .191 (with standard deviation .036). Next, regularization was applied using the Elastic Net. The optimal model (with the smallest expected prediction error) turns out to be the unregularized analysis; if we choose the model that has the smallest expected prediction error within 1 standard deviation, we obtain a sparse model with both POPUL and INCOME omitted from the predictor set, resulting in an expected prediction error of .216(.044). The values for the Ridge and Lasso penalties are 9.00 and .900, respectively. The third option uses spline transformations on the basis of the partial residual plots from the second analysis (see Figure 8); the red curves display for each of the variables the nonlinearity that is present in the results. The third option includes nonmonotonic spline transformations for POPUL, INCOME, and FREEZE with cubic splines, one interior knot (3,1) and a monotonic spline transformation for ILLIT (2,1). First, regularization parameters for the Elastic Net were set to 0.0; compared to the base model (option 1), the r2 obviously increases (.933), but so does the expected prediction error, becoming .210(.045). If we apply the elastic net in addition to the optimal scaling transformations, the selected model has an expected prediction error of .176(.037), being within one standard deviation from the predictor error for the optimal model, which is .159(.030). In the selected model, the predictor POPUL has been omitted from the predictor set, but compared to the first Elastic Net model (option 2), the transformed variable INCOME remains in the model, with corresponding values for the Elastic Net penalties 3.00 for the Ridge penaly and .800 for the Lasso penalty, respectively. The transformations without regularization (not shown) closely resemble the curves in the partial residuals plots from the regularized linear analsis (figure 8). With regularization, the transformations of POPUL, INCOME, and FREEZE change considerably. This can be explained by the role of POPUL. When POPUL drops out of the model, the transformations of INCOME and FREEZE change, and this influences the transformation of POPUL itself. Concluding this example, smaller expected prediction error is obtained when optimal scaling and regularization are applied together, even if the number of parameters that are fitted for the predictors increases from 7 to 14. When optimal scaling is applied, the value of the Ridge penalty drops from 9.00 to 3.00. The DLD drops considerably when optimal scaling is applied without regularization. This illustrates that optimal scaling diminishes multicollinearity among the predictors, hence less Ridge and/or Lasso shrinkage is needed. The drop of DLD is much smaller when optimal scaling is combined with regularization. This can be explained by the role of the regularized coefficients in removing the contribution of the other predictors when estimating a predictor transformation: because the coefficients are shrunken, the contribution of the other predictors is not fully removed. We can depict the effect on DLD by plotting the values for *tolerance*, being the inverse of the diagonal elements of the inverse predictor correlation matrix (Figure 7). It is clear that optimal scaling without regularization OS (upper curve) produces the largest values for *tolerance* when compared to the two other curves at the bottom of the panel (ROS-EN and LIN, respectively), especially for predictor 1,2, and 7 that obtained a nonmonotonic transformation. The Ridge penalty also has its influence. If we fit an additional curve for OS combined with the LASSO, we obtain the value 0.20 for the Lasso penalty, which is slightly higher than the optimal model for ROS-EN (0.0,.10) in Table 8, with EPE=.173(.033). The corresponding curve for the tolerance values is perfectly in between ROS-EN and OS. The corresponding value for DLD equals 3.217.

*Table 7. Data Example 2: Four model options for United States data, with/without Elastic Net regularization and/or optimal scaling. $\lambda_2$=Ridge penalty, $\lambda_1$=Lasso penalty .*

| Transfor-mation | Regula-rization | $r^2$ | EPE optimal | Value $\lambda_2$ | Value $\lambda_1$ | EPE selected | Value $\lambda_2$ | Value $\lambda_1$ | Total # df | DLD |
|---|---|---|---|---|---|---|---|---|---|---|
| 1. No | No | .876 | .191(.036) | - | - | .191(.036) | - | - | 7 | 4.121 |
| 2. No | Yes | .825 | .191(.036) | 0.00 | 0.00 | .216(.044) | 9.00 | .900 | 5 | 4.121 |
| 3. Yes[1] | No | .933 | .210(.045) | - | - | .210(.045) | - | - | 18 | 2.603 |
| 4. Yes[1] | Yes | .865 | .159(.031) | 0.00 | 0.10 | .176(.037) | 3.00 | .800 | 14 | 3.965 |

[1] POPUL, INCOME and FREEZE transformed with cubic nonmonotonic spline, one interior knot, ILLIT transformed with quadratic monotonic spline, one interior knot.

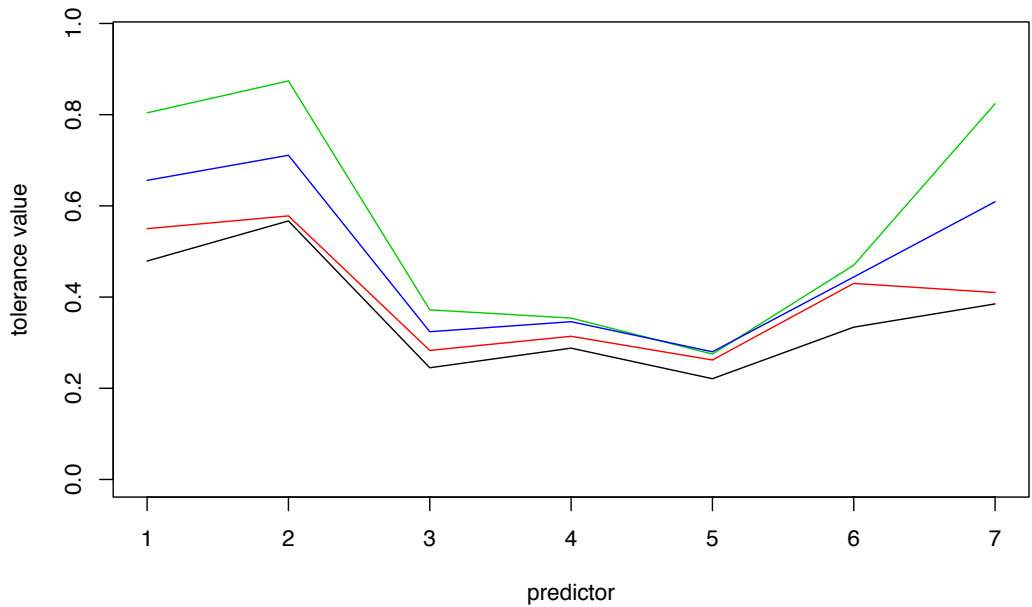

*Figure 7. Tolerance values for the seven predictors in the States Data. The curves correspond (from top to bottom) to: OS regression, ROS regression with LASSO, ROS regression with ENET, and OLS regression.*

In Figure 11, different paths are displayed for Ridge penalties ranging from 0.00 to 3.0, with a stepsize of 1.0. The horizontal axis represents the size of the Lasso penalty, ranging from 0.0 to 1.7, and the vertical axis gives the prediction error, obtained with the .632 bootstrap (EPE). The curve on the bottom gives the EPE for the Lasso penalty 0.0, and shows that the smallest value for EPE is obtained for the Lasso penalty 0.10. From this point, the curve is monotonically increasing. The picture for the three other curves (Lasso penalties from 1.0 up to 3.0) show a completely different picture. Values always are very large for small values for the Lasso penalty, and reach their smallest value at the value 0.80 for the Lasso penalty. The two dots indicate the smallest overall value 0.159(.031) and the smallest value within one standard deviation 0.176(.037) respectively. The latter is on the curve for the Lasso penalty 3.0.

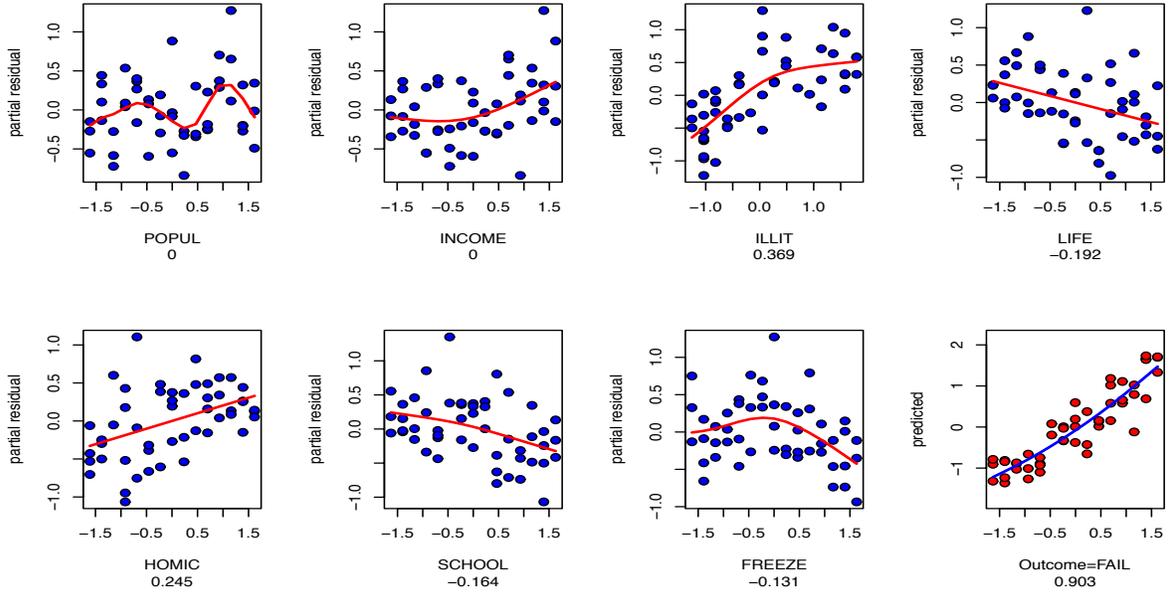

*Figure 8. Partial residuals from linear analysis with regularization (Elastic Net); curves display possible nonlinearity present in the residuals.*

Figure 13 shows all the paths for Ridge penalties ranging from 10 (at the top) to 0.0 (at the bottom). This figure shows that even for very large Ridge penalties, the lowest values for the prediction error are always for Lasso penalties in the range from 0.80 to 1.0. Figure 11 gives the paths for the Apparent Prediction Error that is actually minimized. The path for Ridge penalty 0.0 (starting at the bottom left) now gives a path very close to a straight line for increasing values of the Lasso penalty. The other 10 paths for the Ridge penalty ranging from 1.0 to 10.0 are smoother versions of the ones for the Bootstrap Predicted Error in Figure 12.

## 7. THE GROUP LASSO AND REGULARIZING THE REGRESSION WEIGHT FOR A CATEGORICAL VARIABLE

In standard linear regression, it is common practice to deal with a categorical variable by replacing it by a set of $C_k$ dummy variables $X_k$, where $C_K$ denotes the number of categories. Since each dummy variable $x_{c_k}, c_k = 1, .., C_k$ becomes a binary predictor variable, a coefficient $a_{c_k}$ is sought for each dummy variable. Because columns in $X_k$ are orthogonal, coefficient $a_{c_k}$ is simply found as

$$a_{c_k} = (x'_{c_k} x_{c_k})^{-1} x'_{c_k} (\mathbf{y} - \sum_{l \neq k} \sum_{m=1}^{C_l} a_m x_m) \tag{38}$$

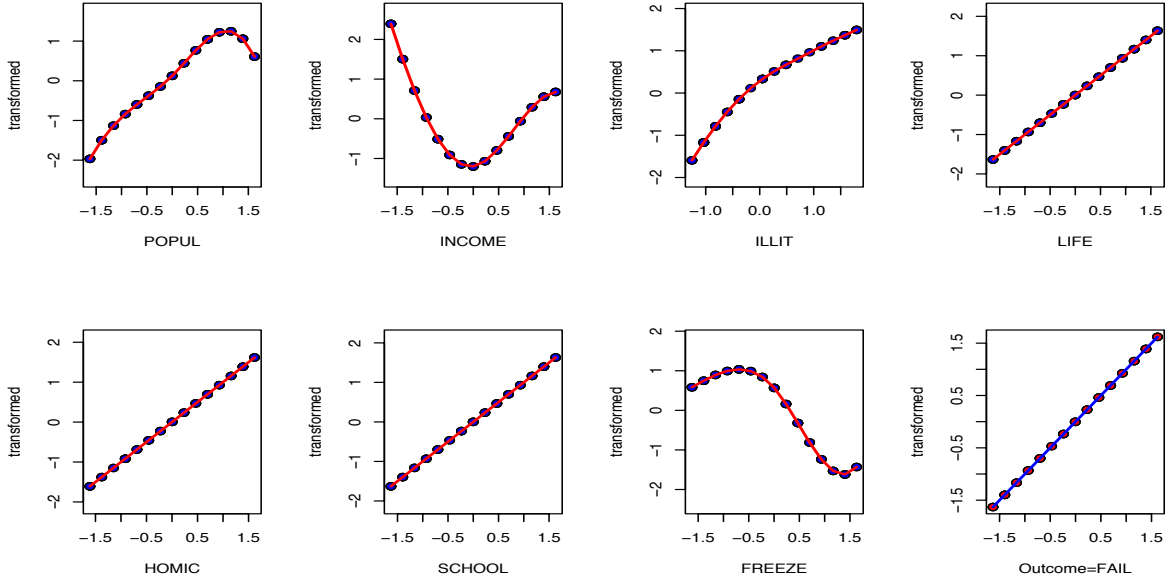

*Figure 9. Optimal scaling transformations from regression with ENET regularization.*

Applying regularization straightforwardly in this situation would amount to regularizing the coefficients $a_{c_k}$ for each dummy variable $x_k$. The Group Lasso method of Yuan and Lin (2006) and the Blockwise Sparse Regression (BSR) method of Kim et al. (2006) treat a dummy variable as a group/block ($X_k$), and apply a norm restriction to the coefficients in the group/block. It can be shown that this restriction is equivalent to applying regularization to the coefficient $\beta_k$ in ROS regression, after rescaling the category quantifications $\mathbf{v}_k$ so that $\mathbf{v}'_k \mathbf{D}_k \mathbf{v}_k = N$. This can be seen by noting that Equation (38) corresponds to the computation of a single category in $\mathbf{v}_k$:

$$\tilde{v}_{c_k} = d_{c_k}^{-1} \mathbf{g}'_{c_k} (\mathbf{y} - \sum_{l \neq k} \beta_l \mathbf{G}_l \mathbf{v}_l), \tag{39}$$

where $\mathbf{g}_{c_k}$, a column of $\mathbf{G}_k$, is equal to $x_{c_k}$ in (38), and $d_{c_k} = \mathbf{g}'_{c_k} \mathbf{g}_{c_k} = (x'_{c_k} x_{c_k})$. So, by collecting the $a_{c_k}$ from (38) in the vector $\tilde{\mathbf{a}}_k$, and rescaling:

$$\mathbf{a}_k = N^{1/2} \tilde{\mathbf{a}}_k (\tilde{\mathbf{a}}'_k \mathbf{D}_k \tilde{\mathbf{a}}_k)^{-1/2}, \tag{40}$$

the coefficients from linear regression on dummy variables yield exactly the same values as the category quantifications in (15). After the coefficients for the dummies are standardized, a regularized regression coefficient associated with the $k$th block of dummies $X_k$ can be computed, and this is equivalent to the solution of the Group Lasso. In analogy with spline coefficients $\mathbf{b}_k$ applied to columns of the spline basis $\mathbf{S}_k$, we could refer to quantifications $\mathbf{y}_c^k$ as coefficients to indicator variables $\mathbf{g}_c^k$, columns of the indicator matrix $\mathbf{G}_k$.

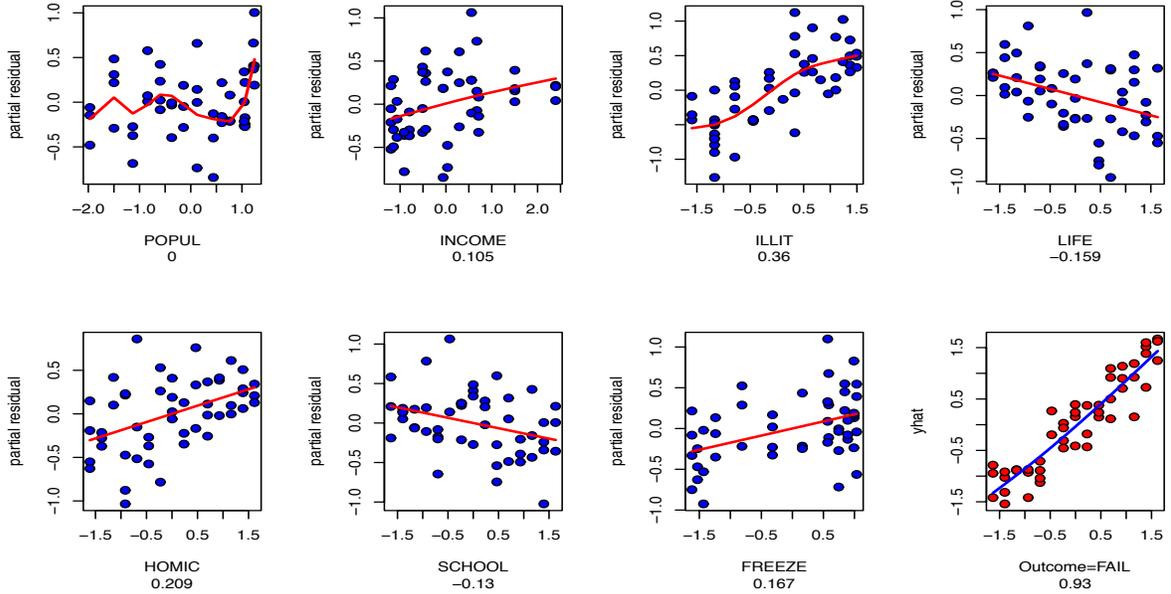

*Figure 10. Partial residuals from Regularized Optimal Scaling regression (ENET).*

For continuous variables, the analogy is similar. In the Group Lasso and the BSR approach, a continuous predictor is represented by a group/block of basis functions, such as polynomials. In the Optimal Scaling approach, continuous predictors are smoothly transformed by applying (non)monotonic regression splines, using an I-spline basis $\mathbf{S}_k$ and fitting spline coefficients $\mathbf{b}_k$ as in (20), and regularization is applied to the associated regression weights $\beta_k$. Concluding, the regularized regression weights associated with the step functions or the regression splines in optimal scaling are equivalent to the results obtained in the Group Lasso and Blockwise Sparse Regression. The regularized optimal scaling approach has the additional advantage that we can also apply *monotonic* step functions and *monotonic* regression splines, as was detailed in section 2. Instead of applying the "Group-Lasso approach" as is automaticaly done in optimal scaling, there are also sometimes advantages of applying regularization to the category quantifications $\mathbf{v}_k$ instead of the regression weights $\beta_k$ (which would be equivalent to regularizing the weights for the dummy variables in standard linear regression). This would be the case when optimal scaling using step functions is applied to categorical variables in which some of the categories have very small marginal frequencies. In those instances, optimal scaling sometimes results in a "degenerate" quantification of the categories, in which only the low-frequency category is distinguished from the other categories, which all receive the same quantification. By applying regularization to the category quantification themselves (instead of to the regression weights), the low-frequency category quantification might shrink to zero (using the standard Lasso penalty).

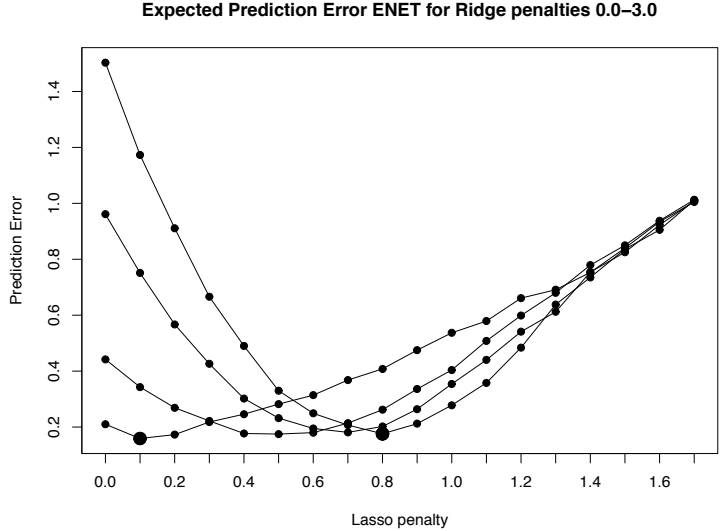

*Figure 11. Expected Prediction Error obtained for Elastic Net Regularized Optimal Scaling Regresssion. Paths from top to bottom represent decreasing values for the Ridge parameter from 3.0 to 0.0. Optimal value found for Ridge penalty 0.0; 1 standard error rule gives Ridge penalty 3.0 .*

In the same spirit, regularization of individual coefficients is applied in our OS approach when we apply monotonic spline transformations. To obtain a spline transformation, a predictor $\mathbf{x}_k$ is represented by a spline basis $\mathbf{S}_k$, set up according to the degree of the polynomial and the number of interior knots, and the associated transformation is written as $\varphi_k(\mathbf{x}_k) = \mathbf{S}_k \mathbf{b}_k = \sum_{t=1}^{T_k} b_t^k \mathbf{s}_t^k$. When $\varphi_k(\mathbf{x}_k)$ in (22) is required to be monotonic with $\mathbf{x}_k$, some of the $b_t^k$ are possibly set to zero to satisfy the monotonicity constraints (that require that all $b_t^k \geq 0$). This is a form of regularization that we could regard as *hard* regularization, and instead, we could also consider *soft* regularization, where we shrink the spline coefficients slowly. We will address this idea somewhat further in the discussion.

## 8. DISCUSSION

In this paper we integrated Optimal Scaling regression with popular regularization methods (Ridge Regression, Lasso, and Elastic Net) in a very general algorithm that can deal with both continuous and categorical variables. Categorical predictors may have either ordered (ordinal) or unordered (nominal) values. The same applies to the outcome variable, although we would prefer to view a categorical outcome variable with unordered categories as requiring a different technique such as discriminant analysis or logistic regression. The need for optimal scaling has various, different reasons. First, the presence of categorical variables calls for quantification, and these may be nominal or ordinal. Second, transformation of continuous variables is called for when nonlinear relationships exist between predictor variables and the outcome. Optimal Scaling linearizes these relationships, as can be seen from the partial residual plots. Third, OS is beneficial when predictor variables are highly correlated, and when a predictor itself can be predicted from the other predictors.

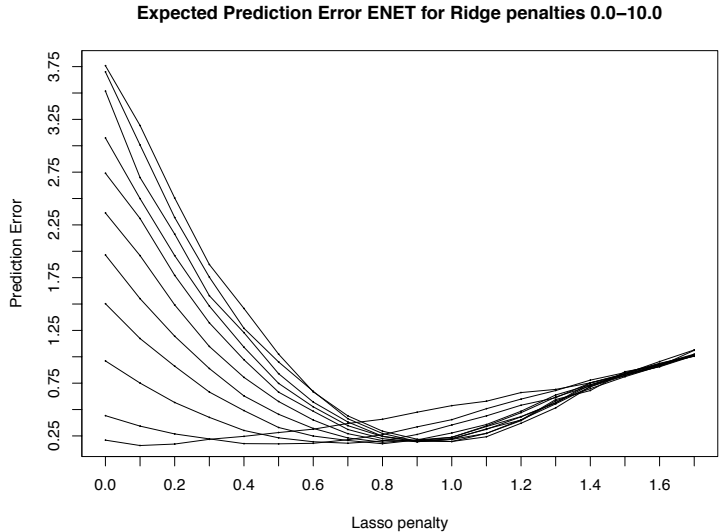

*Figure 12. Expected Prediction Error obtained by Elastic Net Optimal Scaling Regresssion. Paths from top to bottom represent decreasing values for the Ridge parameter from 10.0-0.0.*

The LASSO has a very exciting history (e.g., see Tibshirani 2011). It was already shown in Van der Kooij (2007) that the alternating least squares (ALS) approach that needs to be applied in OS to find the optimal transformations and regression weights (one variable at a time), automatically leads to very simple and efficient estimates for regularized regression coefficients in the LASSO and the Elastic Net. We may conclude that the ALS framework that was kept alive all these years in optimal scaling, gave rise to renewed interest and exciting research using coordinate descent optimization (for example, see Friedman, Hastie, and Tibshirani (2010, 2012), Mazumder, Friedman, and Hastie (2011), Chouldechova and Hastie (2015), to mention only a few references).

The ALS framework computes quantifications for the categories of a predictor as weighted averages of partial residuals (14), based on transformations of the other predictors (9). The approach never creates dummy variables. The straightforward regularized version of Optimal Scaling automatically gives us results equivalent to the Group Lasso and Blockwise Sparse Regression.

In the context of a regularized analysis, there are two goals: model selection and assessment of the selected model. To achieve these goals, the best approach is a *three-way* data split, dividing the data into a training set, a validation set, and a test set. The training set is used for model fitting and the prediction error for model selection is estimated using the validation test. In the end, the prediction error for the selected model (the generalization error) is estimated by applying the model to the test set. This was done in the analysis of the Cervix Cancer Data in Section 3. When there are not enough data for a three-way split, the validation step is approximated either

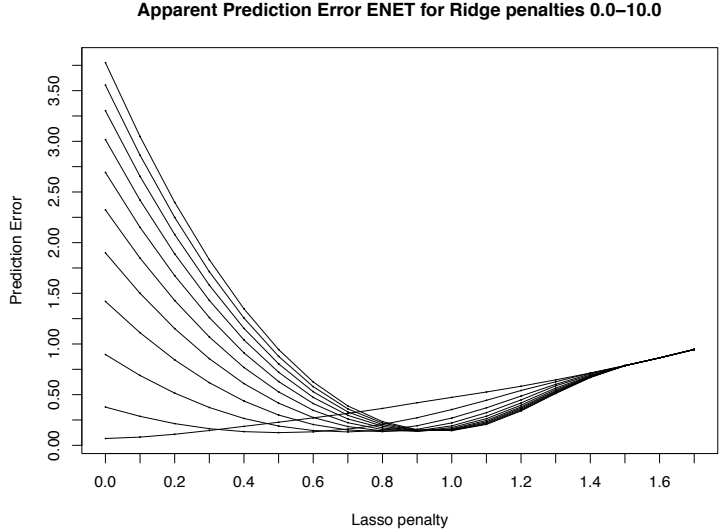

**Apparent Prediction Error ENET for Ridge penalties 0.0–10.0**

*Figure 13. Apparent Prediction Error obtained by Elastic Net Regularized Optimal Scaling Regression. Each path shows values for the Lasso penalty. Different paths represent decreasing values for the Ridge penalty, from 10.0 (top) to 0.0 (bottom).*

analytically, with Generalized Cross Validation (GCV; Golub et al. (1979), AIC, or BIC, or by using a resampling technique, such as cross validation. Of course, the results for the expected prediction error will be too optimistic if we use cross-validation to estimate both the regularization parameters and the prediction error. However, if we are mainly interested in comparing the prediction error for different models, such as with/without regularization and/or optimal scaling, this does not effect the conclusions.

In many analyses that we have seen, some of which are presented in this paper, optimal scaling diminishes the need for strong regularization in case of multicollinearity. If the predictor correlation matrix is ill-conditioned, optimal scaling improves upon this condition, as measured by the value of the smallest eigenvalue. We also proposed the Divergence of Log Determinants $\mathbf{D}_{\ell d}(\mathbf{R}, \mathbf{I})$ to quantify the conditionality of the predictor correlation matrix in a single diagnostic. As for the predictors, Optimal Scaling tends to increase their conditional independence (on average), as measured by so-called tolerance values (given by the inverse of the diagonal elements of the inverse of the correlation matrix).

There is much room left for regularization in Optimal Scaling regression; for instance, we can modify the type of constraints involved in monotonic spline transformation. When we use basis splines $\mathbf{s}_t^k$ as predictors, we could regularize the spline coefficients $b_t^k$, starting with many basis splines representing nonlinear transformations with many degrees of freedom, and restricting these in the iteration process by shrinking the spline coefficients, possibly to zero. When applied to negative coefficients, this automatically will give monotonic transformations. When quantifying categorical predictors, we could set bounds for individual category quantifications $y_c^k$, for e.g. to restrict their absolute values not to be greater than three standard errors from the mean.

In some cases, large Ridge and/or Lasso penalties may be required to prevent overfitting when allowing for optimal transformations. In other instances, the use of optimal scaling may reduce the size of the penalties. This situation calls for further research requiring carefully set up simulation studies to explain these effects.

## NOTE

The algorithm described in this paper has been implemented in a procedure called CATREG that has been developed by the authors in the CATEGORIES module of IBM/SPSS Statistics. Variables are assumed to be categorical (hence the name CATREG), but a straightforward way to allow continuous variables in the analysis is provided by an internal procedure that digitizes continuous data by a linear transformation. Regularization using Ridge regression, the Lasso, and the Elastic Net were included for the first time in version 17.0, as well as Model Selection using cross-validation and the .632 bootstrap. Model Testing is included by the use of so-called supplementary cases.

## ACKNOWLEDGEMENTS

The authors would like to thank Bradley Efron and Jerome Friedman for their suggestions and comments on a previous version of this paper.